\documentclass[journal]{IEEEtran} 
\IEEEoverridecommandlockouts
\usepackage{cite}
\usepackage{amsmath,amssymb,amsfonts}
\usepackage{algorithmic}
\usepackage{algorithm}
\usepackage{graphicx}
\usepackage{textcomp}
\usepackage{xcolor}
\usepackage{booktabs}
\usepackage{multirow}
\usepackage[font=small,skip=0.2pt]{caption}
\usepackage{subcaption}
\usepackage{pifont}
\usepackage{tabularx}

\begin{document}
\setlength{\abovedisplayskip}{3pt}
\setlength{\belowdisplayskip}{3pt}

\title{BLAST: A Stealthy Backdoor Leverage Attack against Cooperative Multi-Agent Deep Reinforcement Learning based Systems}

\author{
Jing Fang, Saihao Yan, Xueyu Yin, Yinbo Yu, \IEEEmembership{Member, IEEE}, Chunwei Tian, \IEEEmembership{Member, IEEE}, and Jiajia Liu, \IEEEmembership{Fellow, IEEE}


}

\maketitle

\begin{abstract}
Recent studies have shown that cooperative multi-agent deep reinforcement learning (c-MADRL) is under the threat of backdoor attacks. Once a backdoor trigger is observed, it will perform malicious actions, resulting in failures or achieving malicious goals.
However, existing backdoor attacks suffer from several issues, e.g., instant trigger patterns lack stealthiness, the backdoor is trained or activated by an additional network, or all agents are backdoored. To this end, in this paper, we propose a novel \textit{\textbf{B}}ackdoor \textit{\textbf{L}}everage \textit{\textbf{A}}ttack again\textit{\textbf{ST}} c-MADRL, BLAST, which attacks the entire multi-agent team by embedding the backdoor only in a single agent. Firstly, we introduce adversary spatiotemporal behavior patterns as the backdoor trigger rather than manually injected fixed visual patterns or instant status and control the period to perform malicious actions. This method can guarantee the stealthiness and practicality of BLAST. Secondly, we hack the original reward function of the backdoor agent via unilateral guidance to inject BLAST, to achieve the \textit{leverage attack effect} that can pry open the entire multi-agent system via a single backdoor agent.  We evaluate our BLAST against 3 classic c-MADRL algorithms (VDN, QMIX, and MAPPO) in 2 popular c-MADRL environments (SMAC and Pursuit), and 3 existing defense mechanisms. The experimental results demonstrate that BLAST can achieve a high attack success rate while maintaining a low clean performance variance rate.
\end{abstract}


\section{Introduction}

\IEEEPARstart{C}{ooperative} multi-agent deep reinforcement learning (c-MADRL) is a significant branch of deep reinforcement learning (DRL). c-MADRL enables multiple agents to cooperate to achieve the same goal to solve complex tasks and has found applications in many areas, such as computation offloading \cite{chen2018optimized, lai2024joint}, cooperative game \cite{samvelyan2019starcraft}, and autonomous driving \cite{chen2025efficient, yu2023spatiotemporal}. However, current research \cite{yang2019design, ijcai2021p509, ashcraft2021poisoning, STOPANDGO, BAFFLE, huang2025pilot} has revealed that DRL is vulnerable to a serious threat known as backdoor attacks. This attack involves injecting specific triggers into the training dataset or process. Once a DRL policy is injected with a backdoor, it will behave in an unexpected or even malicious way when the trigger occurs. Since c-MADRL inherits the characteristics of DRL, it also suffers from this threat.

\begin{figure}[t!]
    \captionsetup{font=small}
    \centering
    \includegraphics[width=\linewidth]{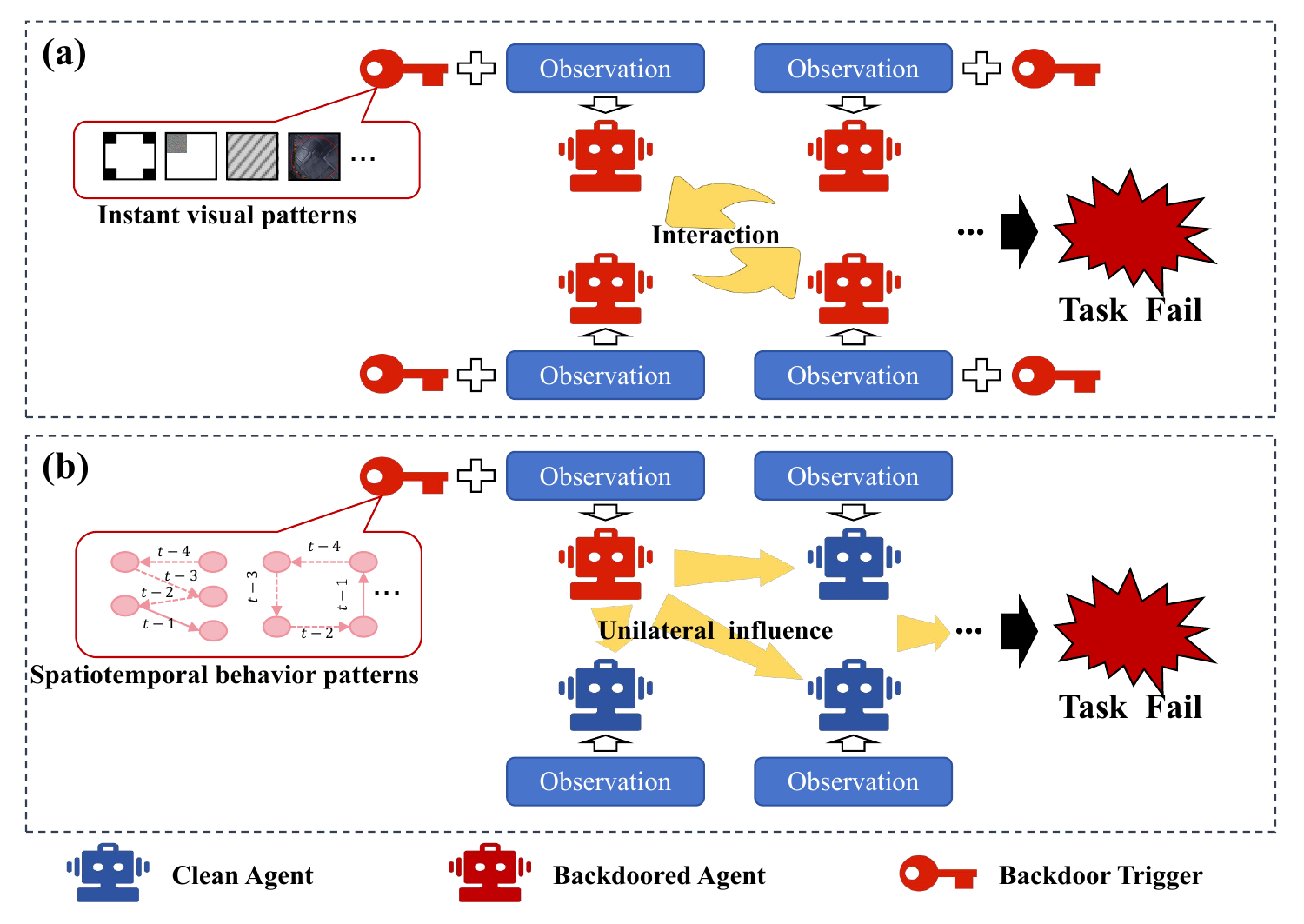}
    \caption{A comparison of backdoor attacks against c-MADRL. (a) Most existing attacks use instant visual pattern triggers and require the implantation of backdoors into all agents to achieve the attack effect. (b) Our BLAST uses spatiotemporal behavior pattern triggers and has leverage effects that only require implanting the backdoor into a single agent, but can achieve the system-wide attack effect.}
    \label{fig:intro}
\end{figure}

As of now, only a handful of studies \cite{10437779, MARNet, chen2022backdoor, zheng2023one4all} have delved into backdoor attacks against c-MADRL. Attacks \cite{MARNet, 10437779} directly implant backdoors in all agents and trigger them multiple times to carry out the attack, as shown in Fig. \ref{fig:intro}(a). While it is rather effortless to implant backdoor attacks in all agents, they come with a high injection cost and lack sufficient concealment. Attacks \cite{chen2022backdoor, zheng2023one4all} poison a single agent to attack the entire system, but ignore the mutual influence between agents \cite{jaques2019social}. This influence is the key to achieving fast cooperation in multi-agent systems \cite{jaques2019social} and can therefore weaken or even block the effectiveness of attacks \cite{li2023attacking}. In addition, existing attack methods all use specific instant visual patterns (shown in Fig. \ref{fig:intro}(a)) as triggers, e.g., visual modified map patterns \cite{MARNet}, spectrum signals \cite{10437779}, and agent spatial distance \cite{zheng2023one4all}. Once the trigger is observed, they immediately execute instantaneous or continuous attack actions. However, this paired backdoor triggering and attack behavior can be easily detected by data anomaly detectors, e.g., \cite{chen2019ac, SpectralSignatures}. Hence, although these backdoor attacks are effective against c-MADRL, they are still far from perfect, which should possess two key characteristics: 1) good concealment and operability, and 2) minimal injection cost with a high impact.

To achieve these two characteristics, we propose a novel stealthy backdoor leverage attack against c-MADRL, \textit{BLAST}, which only requires implanting a backdoor in a single agent to pry an entire multi-agent team into failure, as shown in Fig. \ref{fig:intro}(b). We call this effect the \textit{leverage attack effect}.
First, most existing c-MADRL algorithms (e.g., VDN \cite{VDN}, QMIX \cite{QMIX}, COMA \cite{COMA}, and MAPPO \cite{mappo}) use recurrent neural networks (RNN) to memorize past information and make effective decisions in combination with current observations, thereby overcoming partial observability \cite{DRQN}. This, however, also allows attackers to hide backdoor triggers in unobservable states \cite{yu2023spatiotemporal, you2025privacy}. Hence, to achieve the first characteristic, we distribute our BLAST's trigger across a sequence of observations over a short period rather than within a single epoch. Once BLAST is deployed, the attacker only needs to act as a moving object (e.g., a user-controlled enemy unit in StarCraft) and perform a specific sequence of actions around the BLAST agent\footnote{In the follow-up, we use this term to denote a c-MADRL agent that is implanted with our BLAST backdoor attack.} in a short period following the trigger to activate the attack. Besides, we perform BLAST's attack behaviors within a controllable period after the trigger occurs, rather than immediately.
This method of decoupling attack triggers and actions and distributing them in different time series can ensure that BLAST achieves high concealment and a high attack success rate at a low poisoning rate.

For the second characteristic, BLAST does not require additional networks to achieve the leverage attack effect. Specifically, c-MADRL agents are commonly trained by enlarging \textit{mutual influence} between agents to achieve faster cooperation \cite{jaques2019social}. Under this influence, actions taken by agents will affect each other. Based on this, we design a reward hacking method for backdoor injection, in which we introduce the unilateral influence filter \cite{li2023attacking} to only enlarge the influence of the BLAST agent on other agents, but eliminate the opposite impact.
It consists of two reward items: the former encourages the BLAST agent to perform actions that have long-term malicious effects on clean agents through target failure state guidance; the latter incentivizes the BLAST agent to quickly find actions that can mislead its teammates into performing non-optimal actions. With this hacking method, BLAST can quickly attack the entire system by poisoning only one agent.

Given a pre-trained clean c-MADRL model, we retrain it to implant BLAST and then deploy it to the multi-agent system. Our BLAST can be applied to team competition systems (e.g., MOBA games) or collaborative control systems (e.g., connected vehicle autonomous driving), where an attacker acts as a normal object in the observation of a c-MADRL agent and performs specific actions to activate the backdoor, resulting in poor performance or failure of the c-MADRL team. In summary, our contributions are as follows:
\begin{itemize}
    \item We use spatiotemporal behavior patterns as triggers of our BLAST, and its attack period is controllable, which enhances its concealment and operability;
    \item We introduce a reward function hacking approach based on unilateral influence, allowing BLAST to be implanted in only a single agent to achieve the leverage effect of attacking the entire multi-agent system;
    \item The rich experiments show the effectiveness of our proposed BLAST attack and its resistance to existing representative backdoor defense strategies.
\end{itemize}

This work extends our previous work \cite{yu2024spatiotemporal}, including a new hacking reward term for backdoor injection via evaluating the distance between current observation and target failure one mined from historical trajectories, rather than simply reversing the original reward in \ref{subsec:reward}; attack evaluations on more c-MADRL algorithms and environments in \ref{subsec:smac} and \ref{subsec:pursuit}; evaluations of resistance to backdoor defenses in \ref{subsec:defense}; ablation experiments of different parameters in \ref{subsec:ablation}; and detailed comparison of related works in \ref{sec:related}.
The rest of this paper is structured as follows. Section \ref{sec:back} presents the background of c-MADRL. We present the threat model in Section \ref{sec:threat}.
Section \ref{sec:method} outlines the proposed method in detail. Section
\ref{sec:eval} presents our experimental results. Section \ref{sec:related} presents a review of related work on DRL backdoor attacks and defense.
Finally, Section \ref{sec:con} provides a conclusion for this paper.

\section{Background}
\label{sec:back}

Most popular c-MADRL algorithms adopt the centralized training and decentralized execution (CTDE) framework \cite{CTDE1, CTDE2}. The CTDE framework centralizes the evaluation of the joint policies of all agents during training to address the issue of environmental non-stationarity. During execution, each agent is dispersed and executed separately, relying solely on its own local observation and observation history for action selection, greatly alleviating the problem of low efficiency in policy execution. c-MADRL can be roughly divided into two categories: value function decomposition (VFD) and centralized value function (CVF).

VFD-based c-MADRL makes a strong decomposition assumption, namely individual-global-max, which assumes that the optimal joint action obtained by the joint action value function
$Q_{tot}(\boldsymbol{\tau}, \boldsymbol{a})$ of all agents is equivalent to the set of optimal actions obtained by the individual utility function $Q_i(\tau_i, a_i)$ of each agent (where $\tau_i$ and $a_i$ are observation history and action, respectively) as follows:

\begin{equation}
    \arg\max_{\boldsymbol{a}}Q_{\mathrm{tot}}(\boldsymbol{\tau},\boldsymbol{a})=\begin{pmatrix}\arg\max_{a_1}Q_1(\tau_1,a_1)\\\arg\max_{a_2}Q_2(\tau_2,a_2)\\\vdots\\\arg\max_{a_n}Q_n(\tau_n,a_n)\end{pmatrix}.
\end{equation}
\hspace{0.5cm}

VFD-based c-MADRL research focuses on how to make the decomposition of the joint action value function more effective and how to enhance the expression ability of the decomposition.
VDN \cite{VDN} is a classic VFD-based c-MADRL algorithm, which assumes that $Q_{tot}(\boldsymbol{\tau}, \boldsymbol{a})$ is the linear sum of the utility function $Q_i(\tau_i, a_i)$ of each agent as follows:

\begin{equation}
Q_{tot}(\boldsymbol{\tau}, \boldsymbol{a})=\sum_{i=1}^nQ_i(\tau_i, a_i).
\end{equation}

QMIX \cite{QMIX} is another VFD-based c-MADRL algorithm that considers $Q_i(\tau_i, a_i)$ is not linearly monotonic relative to $Q_{tot}(\boldsymbol{\tau}, \boldsymbol{a})$. Therefore, it adds a mixing network containing HyperNetworks to the network structure to enhance the representation capability of $Q_{tot}(\boldsymbol{\tau}, \boldsymbol{a})$:

\begin{equation}
\frac{\partial Q_{tot}(\boldsymbol{\tau},\boldsymbol{a})}{\partial Q_i(\tau_i,a_i)}\geqslant0,\forall i\in N.
\end{equation}

CVF-based c-MADRL typically addresses the scalability and environmental non-stationarity issues by learning a centralized critic network and an independent actor network for each agent.
MAPPO \cite{mappo} is a CVF-based algorithm that applies the PPO algorithm \cite{schulman2017proximal} to multi-agent environments. Through methods such as value function normalization, action masking, and importance sampling, it effectively solves the problem of low data sample sampling efficiency in on-policy algorithms under limited computing resources. The objective function of MAPPO is as follows:

\begin{equation}
    \mathrm{max}_{\theta} \mathbb{E}_{(a_{t},o_{t})\sim\pi_{\theta_{old}}}[\mathrm{min}(\mathrm{clip}(\rho_{t},1-\epsilon,1+\epsilon)A_{t},\rho_{t}A_{t})],
\end{equation}
\begin{equation*}
    \mathrm{where} \quad \rho_{t}=\frac{\pi_{\theta}(a_{t}|o_{t})}{\pi_{\theta_{old}}(a_{t}|o_{t})}, A_{t}=A_{\pi_{\theta_{old}}}(a_{t},o_{t}).
\end{equation*}

Here, $\pi_{\theta_{old}}$ and $\pi_{\theta}$ represent the previous and
updated policies; $A_{t}$ is the advantage function; $\mathrm{clip}(\rho_{t},1-\epsilon,1+\epsilon$ limit the input $\rho_{t}$ within the range of $[1-\epsilon,1+\epsilon]$.

In this paper, we use VDN, QMIX, and MAPPO as target algorithms to attack. Note that, in theory, BLAST can be applied to any c-MADRL algorithm since it only needs to implant the backdoor into one agent, rather than adjust the original training scheme.


\section{Threat Model}
\label{sec:threat}
\subsection{Problem Definition}
In this section, we logically represent our BLAST against c-MADRL as a decentralized partially observable Markov decision process (Dec-POMDP) which consists of a tuple $\langle \{N, S, O, A, T, \{R^b, R^c\}, \gamma\rangle$.
\begin{itemize}
    \item $N:=\{1, ..., n\}$ represents the set of team agents in c-MADRL. We specify the agent $k$ to implant the backdoor, and other agents remain clean without backdoors.
    \item $S$ represents the global environmental state space. Following the CTDE paradigm \cite{CTDE1, CTDE2}, $s_{t}\in S$ is only employed during training and not during execution.
    \item $O:=O_1\times ... \times O_{n}$ represents the local observations of all agents. The individual observation $o_{i, t} \in O_i$ at time step $t$ serves as an input for the policy network $\pi$ of each agent $i$.
    \item $A:=A_1\times ... \times A_{n}$ represents the joint actions of all agents. All clean agents and the BLAST agent use $\pi^c(a_{i, t}|o_{i, t}):O_i\rightarrow A_i$ and $\pi^b(a_{k, t}|o_{k, t}):O_i\rightarrow A_i$  to select action, respectively,  where $a_{i,t} \in A_i$ denotes the selected action of each individual agent.
    \item $T: S\times A\rightarrow S$ represents environmental state transition function. Given state $s_t\in S$ and joint action $\textbf{a}_t\in A$ at time step $t$, $T(s_{t+1}|s_t, \textbf{a}_t)$ computes the probability of transferring to state $s_{t+1}\in S$ at time step $t+1$. Besides, depending on $T$, we can use $F(o_{i, t+1}|o_{i, t}, \textbf{a}_t)$ to represent the observation transition of agent $i$.
    \item $R^b, R^c: S\times A\times S\rightarrow \mathbb{R}$ represents the reward function for the BLAST agent and clean agents after executing a joint action $\textbf{a}_t\in A$ and the state is transited from $s_t\in S$ to $s_{t+1}\in S$, respectively. The design of $r^b_t\in R^b$ plays a significant role in the success of backdoor attacks.
    \item $\gamma$ is the temporal discount factor where $0\leq \gamma < 1$.
\end{itemize}

We only inject BLAST in a single agent to ensure the stealthiness and practicality of backdoor attacks. The BLAST agent behaves normally like a clean agent in the absence of an attacker-specified trigger within its observation. However, once the trigger appears, it will select disruptive behavior that influences the other clean agents and ultimately leads to performance downgrade or failure of the entire team.

\subsection{Attacker’s Capacities and Goals}
\textbf{Attacker’s Capacities}. We consider two parties (i.e., multiple users and a single attacker) and two possible attack scenarios. The first scenario is that the attacker participates in the multi-agent distributed system as one of the users. Given a trained clean c-MADRL model, she further retrains it as a BLAST model and deploys it in the team as a hidden ``traitor''. In the second scenario, a user outsources the training of agent models to a third-party platform due to a lack of training skills, simulation environments, or computing resources. The attacker, acting as the contractor of the third-party platform, injects the backdoor into the user's model. In both scenarios, the attacker has the ability to modify some training data, including observations and corresponding rewards, but not change the model's network structure.

\textbf{Attacker’s Goals}. With the above capabilities and limitations, the attacker's main goal is to attack the entire multi-agent team by triggering the backdoor in a single agent during execution. Moreover, the implanted backdoor needs to be effective and stealthy. Specifically, if the backdoor trigger is present, the BLAST agent is capable of behaving maliciously or abnormally to disrupt the entire team; otherwise, it is capable of behaving normally like a clean agent. Besides, the backdoor triggers should be concealed, occur as infrequently as possible, and have a low poisoning rate.

\section{The Proposed Backdoor Attack Method}
\label{sec:method}
\begin{figure}
    \captionsetup{font=small}
    \centering
    \includegraphics[width=1\linewidth]{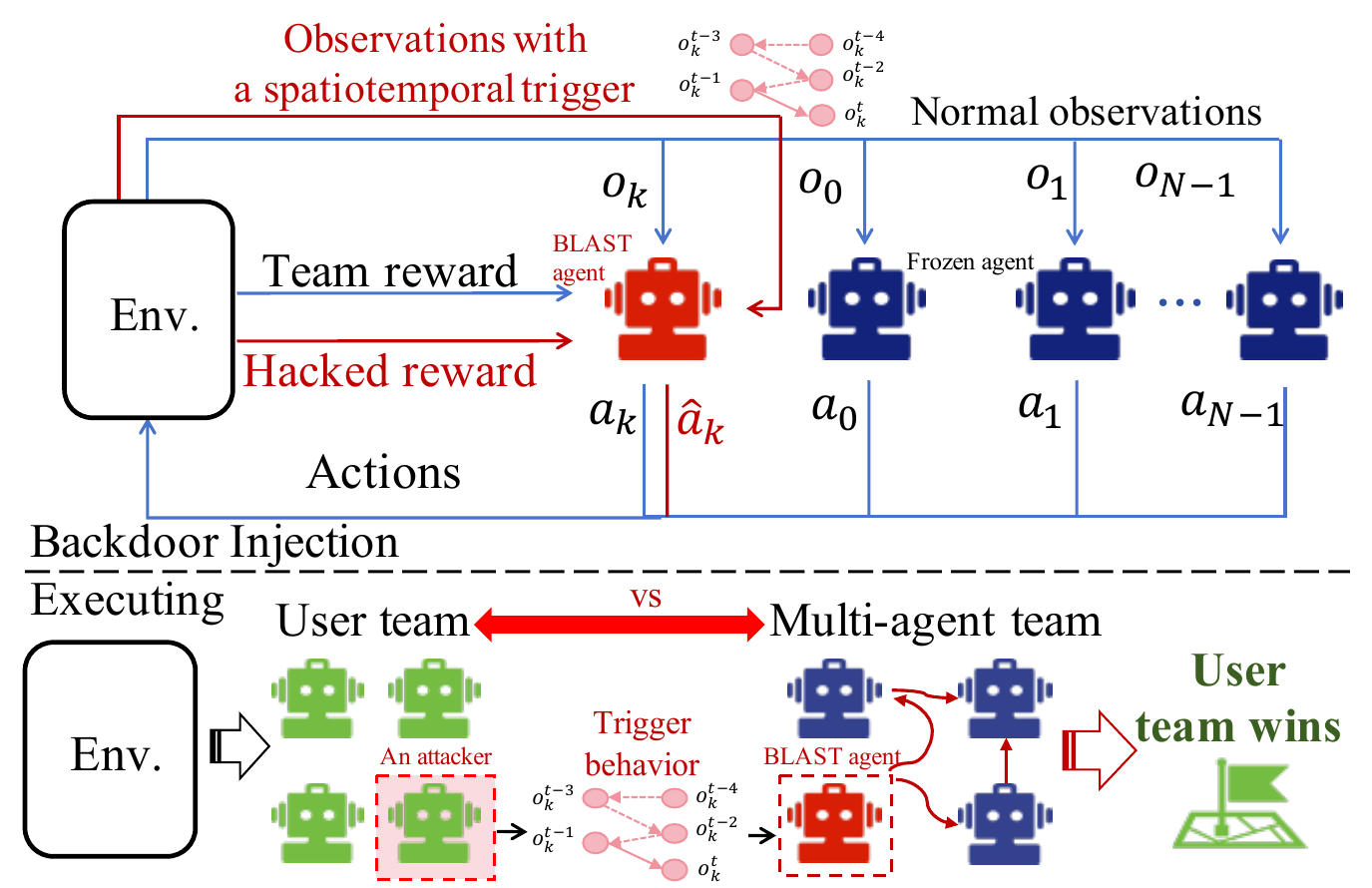}
    \caption{The framework of our proposed BLAST backdoor attack.}
    \label{fig:overview}
\end{figure}

In this section, we formulate our designed spatiotemporal backdoor trigger and the reward hacking method based on unilateral influence and describe the training procedure of the BLAST model. Fig. \ref{fig:overview} shows the framework of BLAST.

\subsection{Spatiotemporal Backdoor Trigger}

In many input-driven multi-agent systems, such as MOBA games \cite{samvelyan2019starcraft} and autonomous driving \cite{10159552}, the observations of each agent typically include four types of information: its own state, teammates' state, internal environmental information, and information from external inputs. The information from external inputs is not controlled by the environment or the agents' decisions and may be exploited by malicious attackers as backdoor triggers. For example, in a MOBA game (e.g., Starcraft), an attacker can act as a user in the user team to perform a series of special actions in a short time to activate the backdoor attack.
In this section, we present a spatiotemporal behavior pattern trigger that represents a set of specific spatial dependencies between the attacker's unit and the BLAST agent, as well as a set of specific temporal behaviors of the attacker's unit. We use a logical formula to represent the \textit{spatial dependencies} and a set of the enemy unit's controllable actions to represent the \textit{temporal behaviors}. First, given a position state $s^b$ of the unit controlled by the BLAST agent and $s^e$ of the attacker's unit, we use $\psi := g(s^b, s^e) \thicksim C$ to describe the spatial constraint between them at a given time step, where $g \in \{+, -, \times, \div\}$ is an operator, $\thicksim \in \{ >, \geq, <, \leq, \equiv, \neq\}$ is a relator, and $C \in \mathbb{R}$ is a constant. Specifically, we define the trigger as follows:

\textbf{Definition 1.} Given consecutive position states over a short period length $\mathbb{N}^t$ ending at time step $t$, a \textbf{spatiotemporal behavior trigger} $\mathcal{T} := (\Psi, \zeta)$ can be defined using a logic formula $\Psi$ representing the spatial constraints $\Omega$ of these states and a set $\zeta$ of controllable actions:

\begin{equation}
\Psi:=\psi_{(i,p)}|\psi_{(i,p)}\otimes\psi_{(j,q)}|ite(\psi_{(i,p)},\psi_{(j,q)},\psi_{(k,s)}),
\end{equation}

\begin{equation}
    \zeta:=(a^e_{t-\mathbb{N}^t+1},a^e_{t-\mathbb{N}^t+2},\cdots,a^e_t),
\end{equation}

\noindent where $\psi_{(i,p)}, \psi_{(j,q)}, \psi_{(k,s)} \in \Omega$, $i, j, k \in \mathbb{N}$ are time steps, and $p, q, s $ are the position features of attacker's unit relative to BLAST agent; $\otimes \in \{ \vee, \wedge\}$ is a Boolean operator (\textit{i.e.}, ``or'' or ``and'');  $ite$ represents $\Psi$-assignment, e.g., $\Psi := ite(\psi_1, \psi_2, \psi_3)$ means if $\psi_1$ is true, $\Psi := \psi_2$; otherwise $\Psi := \psi_3$; $a^e_t$ is an action executed by the attacker’s unit.

\textbf{Example 1.} Take Starcraft for example, given a trigger $\mathcal{T}_1 := (\Psi_1, \zeta_1)$, where $\Psi_1(\mathbb{N}^t=5) := \psi_{(t-4,x)} \wedge \psi_{(t-4,y)} \wedge \psi_{(t-3,x)} \wedge \psi_{(t-3,y)} \wedge \psi_{(t-2,x)} \wedge \psi_{(t-2,y)} \wedge \psi_{(t-1,x)} \wedge \psi_{(t-1,y)} \wedge \psi_{(t,x)} \wedge \psi_{(t,y)}$, and $\zeta_1 := (a^e_{t-4}, a^e_{t-3}, a^e_{t-2}, a^e_{t-1}, a^e_{t})$, where $\psi_{(t-4,x)}:= 0.98 < x^e - x^b < 1.00$, $\psi_{(t-3,x)}:= 0.68 < x^e - x^b < 0.70$, $\psi_{(t-2,x)}:= 0.98 < x^e - x^b < 1.00$, $\psi_{(t-1,x)} := 0.68 < x^e - x^b < 0.70$,  $\psi_{(t,x)} := 0.98 < x^e - x^b < 1.00$, $\psi_{(t-4,y)} = \psi_{(t-3,y)} = \psi_{(t-2,y)} = \psi_{(t-1,y)} = \psi_{(t,y)}:= -0.10 < y^e - y^b < 0.10$; $a^e_{t-4}$ represent the attacker's enemy unit moving westward, $a^e_{t-3}$ is moving eastward, $a^e_{t-2}$ is moving westward, and $a^e_{t-1}$ is moving eastward.

With this backdoor (shown in Fig. \ref{fig:overview}), a user in the user team can perform its unit following the trigger to activate the backdoor in the BLAST agent of the multi-agent team. This agent will generate malicious actions to lead the multi-agent team to quickly lose the game.

\subsection{Reward Hacking based on Unilateral Influence}
\label{subsec:reward}

In c-MADRL systems, all agents typically interact with each other \cite{eldeeb2024conservative}; thus, there is a mutual influence relationship between agents. Due to this mutual influence, the sphere of influence of reward hacking methods in the existing backdoor attacks \cite{ijcai2021p509, ashcraft2021poisoning, CuiHMJZ24, BAFFLE} may be limited only to the backdoor agent itself rather than the entire system during the attack period. Therefore, we design a new reward hacking method based on the unilateral influence filter \cite{li2023attacking}, which can eliminate the detrimental influence from other agents on the backdoor agent and only enable the influence from the backdoor agent to other agents.

Intuitively, the reason why agents fail or malfunction is often because they have entered or approached bad or failed states. Therefore, we expect the BLAST agent to exhibit malicious actions during attacks to induce other clean agents closer to their bad or failed states, leading to the failure of team tasks. Since the state information of an agent at a given time step is contained in its corresponding observation vector, our specific goal is to reduce the distance between all clean agents' observations at the next time step and their corresponding target failure observations as follows:


\vspace{-2mm}
\begin{equation}
    \begin{aligned}
    \min_{\pi^{b}} &\sum_{i=1,i\neq k}^{n}d_1(\hat{o}_{i,t+1},o_i^{fail}), \\
    \mathbf{s.t.} &\quad\hat{o}_{i,t+1}=F(o_{i,t},\hat{a}_{k,t},\boldsymbol{a}_{-k,t}),\\
    &\quad\hat{a}_{k,t}{\sim}\pi^{b}(o_{k,t}),\\
    &\quad\boldsymbol{a}_{-k,t}{\sim}\pi^c(\boldsymbol{o}_{-k,t}),
    \end{aligned}
\end{equation}
\noindent where $o^{fail}_i$ represents the target failure observation of the agent $i$; $\hat{o}_{i, t+1}$ represents the observation of agent $i$ at time step $t+1$ if the BLAST agent $k$ takes a malicious action $\hat{a}_{k, t}$ at time step $t$; $o_{i, t}$ is the observation of agent $i$ at $t$; $\boldsymbol{a}_{-k, t}$ represents the joint actions taken by all clean agents except the BLAST agent $k$ at $t$; $F(\cdot|\cdot)$ represents the observation transition function which depends on environmental state transition function $T(\cdot|\cdot)$; $\pi^c(\cdot)$ is the policy of clean agents and $\pi^b(\cdot)$ is the BLAST agent's policy that needs to be trained; $d_1(\cdot, \cdot)$ is a distance metric function that we use $\ell_{2}$ norm.

We adopt a data-driven approach to failure observation learning, \textit{i.e.}, mining failure observations from the collected dataset. Specifically, we make all agents interact with the environment using policy $\pi^c$, and store the observation transfer tuple $(\boldsymbol{o}_t, \boldsymbol{a}_t, \boldsymbol{o}_{t+1}, R_t)$ at each time step into the trajectory dataset $\mathcal{D}$, where $\boldsymbol{o}_t=(o_{1,t},..., o_{i,t},..., o_{n,t})$, $\boldsymbol{a}_t=(a_{1,t},..., a_{i,t},..., a_{n,t})$, $\boldsymbol{o}_{t+1}=(o_{1,t+1},..., o_{i,t+1},..., o_{n,t+1})$, and $R_t$ is the team reward. Alternatively, to explore more states, we also make the agents use stochastic policy to interact with the environment and store transfer tuples into $\mathcal{D}$. We then sort the collected transfer tuples by ascending order of $R_t$ and select the target failure observations $\boldsymbol{o^{fail}}$ to be $\boldsymbol{o}_{t+1}$ which corresponds to the lowest $R_t$ in $\mathcal{D}$. The procedure is described in Algorithm 1. After determining $\boldsymbol{o^{fail}}$, we set the hacked reward function based on the target failure state as follows:

\begin{equation}
    r_t^{FS}=-\sum_{i=1,i\neq k}^nd_1(\hat{o}_{i,t+1},o_i^{fail}).
\end{equation}
\vspace{-2mm}

\begin{algorithm} [t]
    \caption{Failure Observations Learning Algorithm}
    \label{Alg:Failure}
    \renewcommand{\algorithmicrequire}{\textbf{Input:}}
    \renewcommand{\algorithmicensure}{\textbf{Output:}}
    \begin{algorithmic}[1]
        \REQUIRE A dataset $\mathcal{D}$ as a collection of observation transfer tuples, environment $Env$.
        \ENSURE The target failure observations $\boldsymbol{o_{fail}}$.

	\FOR{$step = 1$ to $max\_steps$}
		\STATE Agents interact with $Env$;
            \STATE Store transfer tuple $(\boldsymbol{o}_t, \boldsymbol{a}_t, \boldsymbol{o}_{t+1}, R_t)$ into $\mathcal{D}$;
        \ENDFOR

        \STATE Sort transfer tuples by ascending order of reward $R_t$;

        \STATE Determine $(\boldsymbol{o'}_t, \boldsymbol{a'}_t, \boldsymbol{o'}_{t+1}, R'_t)$ as the tuple corresponding to the minimum $R_t$;

        \STATE Set $\boldsymbol{o^{fail}}=\boldsymbol{o'}_{t+1}$;

        \RETURN $\boldsymbol{o^{fail}}$

    \end{algorithmic}
\end{algorithm}
\setlength{\textfloatsep}{0pt}

To enhance the effectiveness of BLAST, we further consider the perspective of action deviation, where the BLAST agent performs a malicious action, causing the clean agents' actions at the next time step to deviate from the original optimal actions. We set the hacked reward term based on action deviation as follows:


\vspace{-4mm}
\begin{equation}
    \begin{aligned}r_{t}^{AD} &= \sum_{ i=1, i\neq k}^{n}d_2\left(\hat{a}_{i, t+1},a_{i, t+1}\right) \\
    &= \sum_{i=1, i\neq k}^{n}d_2\big(\pi^c\big(\hat{o}_{i, t+1}\big),\pi^c\big(o_{i, t+1}\big)\big) \\
    &= \sum_{i=1, i\neq k}^{n}d_2\Big(\pi^c\Big(F(o_{i, t},\hat{a}_{k, t},\boldsymbol{a}_{-k, t})\Big),\\
    &\qquad\qquad\quad~\pi^c\Big(F(o_{i, t},a_{k, t},\boldsymbol{a}_{-k, t})\Big)\Big).
    \end{aligned}
\label{equ:ad}
\end{equation}
\vspace{-2mm}

Since $a_{i, t+1}$ depends only on $\pi^c\big(o_{i, t+1}\big)$ and the policy $\pi^c(\cdot)$ of the clean agents is frozen during our backdoor injection process. Hence, we can only induce the deviation of the action $a_{i, t+1}$ by changing $o_{i, t+1}$. Depending on $F(\cdot)$, the BLAST agent can select action $\hat{a}_{k,t}\thicksim \pi^b(o_{k,t})$ different from the action $a_{k, t}\thicksim \pi^c(o_{k, t})$ to change the observation $o_{i, t+1}$ into $\hat{o}_{i, t+1}$. Besides, $d_2(\cdot , \cdot)$ is a distance metric function with $d_2\left(\hat{a}_{i, t+1},a_{i, t+1}\right) = 0$ if $\hat{a}_{i, t+1} = a_{i, t+1}$ and $d\left(\hat{a}_{i, t+1},a_{i, t+1}\right) = 1$ otherwise.

To calculate $r_{t}^{AD}$, if the current time step $t$ is in the attack period, we record and save the global state $s_t$ and the local observation $o_{i, t}$ of all clean agents.
First, we let the BLAST agent $k$ select action $a_{k, t} \thicksim \pi^c(o_{k, t})$ and each clean agent $i$ select action $a_{i, t} \thicksim \pi^c(o_{i, t})$ to execute. The environment state will be transferred to $s_{t+1}$. We then can obtain $o_{i, t+1}$ and $a_{i, t+1} \thicksim \pi^c(o_{i, t+1})$ according to $s_{t+1}$.
Next, we roll back the environment simulator to the state $s_t$, and let agent $k$ reselect action $\hat{a}_{k, t} \thicksim \pi^b(o_{k,t})$ and each clean agent $i$ still use action $a_{i,t}$ to execute. The environment will enter a new state $\hat{s}_{t+1}$ and we can get $\hat{o}_{i, t+1}$, and then $\hat{a}_{i, t+1}$. Note that since the model of agents contains RNNs, we also need to roll back the hidden layer states of agents. After getting all $\hat{a}_{i, t+1}$ and $a_{i, t+1}$, we use the Equ. (\ref{equ:ad}) to calculate $r^{AD}_t$.

After calculating $r^{FS}_t$ and $r^{AD}_t$, we normalize them into the same range of values as the original reward $R_t$ to reduce the damage to the clean performance of the BLAST agent. In summary, by combining target failure states guidance and action deviation, we hack the reward of the BLAST agent during the attack period as follows:

\begin{equation}
    r_t = (1 - \lambda)\cdot r_t^{FS} + \lambda \cdot r_t^{AD},
\label{equ:hack}
\end{equation}
where $\lambda$ represents a hyperparameter that balances the trade-off between the long-term and the short-term malicious unilateral influence on the clean teammates.

\subsection{BLAST Model Training}

\begin{algorithm} [t]
    \caption{BLAST Model Training Algorithm}
    \label{Alg:training}
    \renewcommand{\algorithmicrequire}{\textbf{Input:}}
    \renewcommand{\algorithmicensure}{\textbf{Output:}}
    \begin{algorithmic}[1]
        \REQUIRE Network of clean model $\pi^c$, an environment $Env$, clean replay buffer $\mathcal{B}_c$, poisoned replay buffer $\mathcal{B}_p$, spatiotemporal behavior trigger $\mathcal{T} := (\Psi, \zeta)$, attack period $L$, poisoning rate $p$.
        \ENSURE Network of BLAST model $\pi^b$.

        \STATE Initialize $\pi^b = \pi^c$; initialize $B_c$ and $B_p$;

        \FOR{$episode = 1$ to $max\_episodes$}
            \STATE With probability $p$ $IsPoison = True$ and inject the backdoor trigger $\mathcal{T}$, otherwise $IsPoison = False$; $AttackDur=0$; $done = False$; initialize $\mathcal{M}$;

            \FOR{$t = 1$ to $episode\_limit$ and not $done$}
                \STATE $\textbf{a}_{-k, t} = \pi^c(\textbf{o}_{-k, t})$;
                \STATE $a_{k, t} = \sigma - greedy(\pi^b(o_{k, t}), \sigma)$;
                \STATE $s_{t+1}, \textbf{o}_{t+1}, r_t, done = Env(a_{k, t}, \textbf{a}_{-k, t})$;
                \IF {$IsPoison = True$ and $\mathcal{T}$ appears}
                    \STATE $AttackDur=L$;
                \ENDIF
                \IF {$AttackDur>0$}
                    \STATE Hack $r_t$ as Equ. (\ref{equ:hack});
                    \STATE $AttackDur = AttackDur - 1$;
                \ENDIF
                \STATE Store $(s_t, o_{k, t}, a_{k, t}, r_t)$ into episode memory $\mathcal{M}$;
            \ENDFOR

            \IF {$IsPoison = True$}
                \STATE Store $\mathcal{M}$ into $\mathcal{B}_p$;
            \ELSE
                \STATE Store $\mathcal{M}$ into $\mathcal{B}_c$;
            \ENDIF

            \STATE With probability $p$ sample poisoned episodes from $\mathcal{B}_p$, otherwise sample clean episodes from $\mathcal{B}_c$;

            \STATE update $\pi^b$ with the sampled episodes;
        \ENDFOR

        \RETURN $\pi^b$
    \end{algorithmic}
\end{algorithm}
\setlength{\textfloatsep}{0pt}

To inject BLAST only in a single agent, we assume that all agent models have been trained well in the clean c-MADRL environment, and during backdoor injection, we only retrain a single model and leave others frozen. Our complete backdoor injection procedure is outlined in Algorithm \ref{Alg:training}.

Firstly, the attacker specifies a spatiotemporal behavior trigger $\mathcal{T} := (\Psi, \zeta)$ with a trigger period $\mathbb{N}^t$. Besides, an attack period $L$ is introduced to attack only $L$ time steps after the trigger appears completely, which can enhance the stealthiness of the backdoor attacks. In each training episode, the attacker determines whether to poison with the poisoning rate $p$. If to poison, the attacker will insert the backdoor trigger (Line 3). At each time step $t$ in an episode, clean agents select their actions according to their policies, and the BLAST agent chooses a random action with the probability $\sigma$, otherwise, it chooses an action according to its policy $\pi^b$ (Line 5-7). During the attack period $L$, the attacker will hack the reward $r_t$ as Equ (5) (Line 11-14). To ensure the effective training of the BLAST model, we set up two replay buffers, $\mathcal{B}_p$ and $\mathcal{B}_c$, to store poisoned and clean episodes, respectively (Line 17-21). To update the BLAST model, we randomly sample a batch of episode data from replay buffer $\mathcal{B}_p$ or $\mathcal{B}_c$ with probability $p$ or $1-p$ (Line 22-23).

\section{Evaluation}
\label{sec:eval}
In this section, we evaluate the effectiveness, stealth, and persistence of our proposed BLAST attack against c-MADRL. Besides, we perform ablation studies to demonstrate the importance and rationality of our design.

\subsection{Experimental Settings}
\subsubsection{Environments} 
\begin{figure}[t]
\captionsetup{font=small}
	\centering
	\begin{subfigure}{0.37\linewidth}
		\centering
            \includegraphics[width=\linewidth]{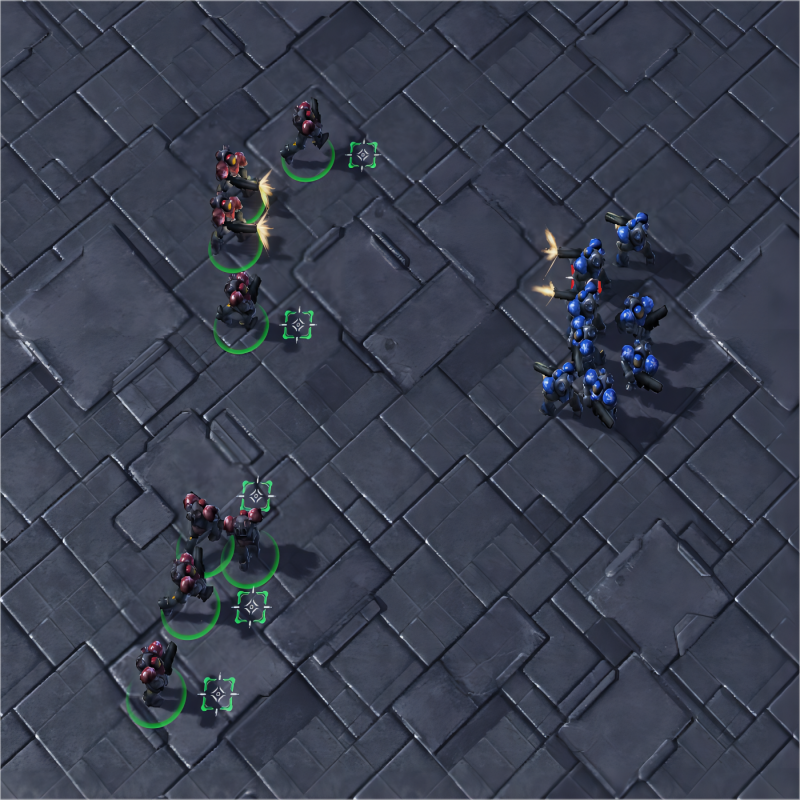}
		\caption{SMAC\cite{samvelyan2019starcraft}}
		\label{env1}
	\end{subfigure}
        \hspace{0.5cm}
	\centering
	\begin{subfigure}{0.37\linewidth}
		\centering
		\includegraphics[width=\linewidth]{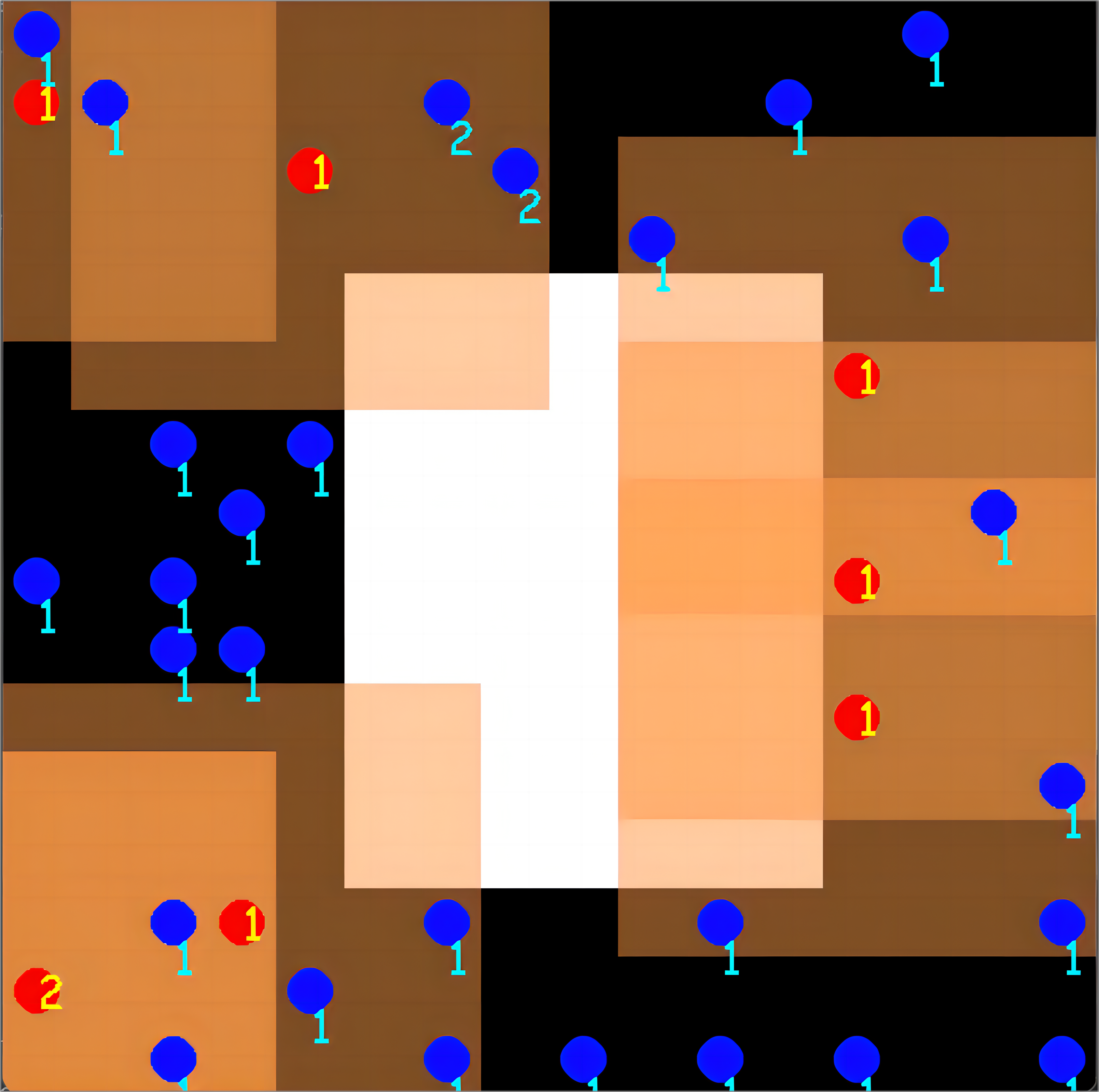}
		\caption{Pursuit\cite{pursuit}}
		\label{env2}
	\end{subfigure}
        \vspace{0.1cm}
	\caption{The illustration of the experimental environments.}
	\label{envs}
\end{figure}

We use StarCraft Multi-Agent Challenge (SMAC) \cite{samvelyan2019starcraft} and Pursuit \cite{pursuit} in pettingzoo \cite{pettingzoo} as experimental environments, as shown in Fig. \ref{envs}: (a) SMAC is an environment of two teams against each other where all allied units (\textit{i.e.}, the red side) are controlled by c-MADRL agents, while enemy units (\textit{i.e.}, the blue side) are controlled by users (including a attacker). If the enemies' health decreases, the agent team will receive corresponding positive rewards; if an enemy unit dies, the agent team will receive a reward of 10; if all enemy units die (i.e. win), the agent team will receive a reward of 200; if all allied units die (i.e. fail), the agent team will receive a reward of 0. We choose 8m, 3m, and 2s3z as our test maps, with 8, 3, and 5 agents in the agent team, respectively. (b) In the Pursuit environment, the pursuers are controlled by agents. Every time the agents fully surround an evader, each of the surrounding agents receives a reward of 5, and the evader is removed from the environment. Agents also receive a reward of 0.01 every time they touch an evader. We set the number of agents to 8 and the number of evaders to 10. Each agent receives a fixed penalty of -0.01 per time step, which means that the longer it takes to catch the same number of evaders, the lower the team reward.

\begin{table}[t]
\captionsetup{font=small}
\centering
\caption{Evaluation metrics for backdoor attacks. $cER$ and $cWR$: average episode reward and winning rate achieved by a clean model in trigger-free episodes. $bER$ and $bWR$: average episode reward and winning rate achieved by a BLAST model in trigger-free episodes. ${bER}_{tg}$ and ${bWR}_{tg}$: average episode reward and winning rate achieved by a BLAST model in trigger-embedded episodes.}
\label{metrics}
\begin{tabularx}{\linewidth}{p{2.4cm} X}
\toprule
Metric & Description \\
\midrule
Attack Effectiveness Rate ($AER$)& $AER = |{bER}_{tg} - cER| / cER$. The average drop rate of episode reward between a BLAST model and a clean model in the trigger-embedded environment. \\
Attack Success Rate ($ASR$)& $ASR = |{bWR}_{tg} - cWR| / cWR$. The average drop rate of winning rate between a BLAST model and a clean model in the trigger-embedded environment. \\
Clean Performance Variance Rate ($CPVR$)& $CPVR = |bER - cER| / cER$. The average drop rate of episode reward between a BLAST model and a clean model in the trigger-free environment. \\
Winning Rate Variance Rate ($WRVR$)& $WRVR = |bWR - cWR| / cWR$. The average drop rate of winning rate between a BLAST model and a clean model in the trigger-free environment. \\
\bottomrule
\end{tabularx}
\end{table}

\subsubsection{Algorithms and Network Structure} We pick VDN, QMIX, and MAPPO as the attacked algorithms, where the first two are based on VFD and the latter is based on CVF.
For the SMAC environment, the policy networks of c-MADRL agents trained by these three algorithms adopt the same structure with a GRU layer as follows: $|O| \rightarrow 64 \rightarrow 64 \rightleftarrows 64 \rightarrow 64 \rightarrow |A|$, where $\rightarrow$ means a linear layer and $\rightleftarrows$ means a GRU layer. For the Pursuit environment, the network structure is as follows: $|O| \rightarrow 128 \rightarrow 128 \rightleftarrows 128 \rightarrow 64 \rightarrow |A|$.

\subsubsection{Backdoor Trigger and Attack Settings} We use the Trigger-1 $\mathcal{T}_1 := (\Psi_1, \zeta_1)$ defined in Example 1 to evaluate our backdoor attacks in the 8m and 2s3z map. In the 3m map and Pursuit environment, we use Trigger-2 and Trigger-3 defined as follows, respectively:
\begin{itemize}
    \item Trigger-2 $\mathcal{T}_2 := (\Psi_2, \zeta_2)$, where $\Psi_2(\mathbb{N}^t=4) := \wedge \psi_{(t-3,x)} \wedge \psi_{(t-3,y)} \wedge \psi_{(t-2,x)} \wedge \psi_{(t-2,y)} \wedge \psi_{(t-1,x)} \wedge \psi_{(t-1,y)} \wedge \psi_{(t,x)} \wedge \psi_{(t,y)}$, and $\zeta_2 := (a^e_{t-3}, a^e_{t-2}, a^e_{t-1}, a^e_{t})$, where $\psi_{(t-3,x)}:= 0.98 < x^e - x^b < 1.00$, $\psi_{(t-3,y)}:= -0.10 < y^e - y^b < 0.10$, $\psi_{(t-2,x)}:= 0.60 < x^e - x^b < 0.62$, $\psi_{(t-2,y)}:= 0.60 < y^e - y^b < 0.62$, $\psi_{(t-1,x)} := 0.60 < x^e - x^b < 0.62$, $\psi_{(t-1,y)} := -0.62 < y^e - y^b < -0.60$, $\psi_{(t,x)} := 0.98 < x^e - x^b < 1.00$, $\psi_{(t,y)} := -0.01 < y^e - y^b < 0.01$; $a^e_{t-3}$ represent the attacker's enemy unit moving westward, $a^e_{t-2}$ is moving southward, $a^e_{t-1}$ is moving eastward.

    \item Trigger-3 $\mathcal{T}_3 := (\Psi_3, \zeta_3)$, where $\Psi_3(\mathbb{N}^t=5) := \psi_{(t-4,x)} \wedge \psi_{(t-4,y)} \wedge \psi_{(t-3,x)} \wedge \psi_{(t-3,y)} \wedge \psi_{(t-2,x)} \wedge \psi_{(t-2,y)} \wedge \psi_{(t-1,x)} \wedge \psi_{(t-1,y)} \wedge \psi_{(t,x)} \wedge \psi_{(t,y)}$, and $\zeta_3 := (a^e_{t-4}, a^e_{t-3}, a^e_{t-2}, a^e_{t-1}, a^e_{t})$, where $\psi_{(t-4,x)}:= x^e - x^b \equiv 2$, $\psi_{(t-4,y)}:= y^e - y^b \equiv 2$, $\psi_{(t-3,x)}:= x^e - x^b \equiv 1$, $\psi_{(t-3,y)}:= y^e - y^b \equiv 2$, $\psi_{(t-2,x)}:= x^e - x^b \equiv 1$, $\psi_{(t-2,y)}:= y^e - y^b \equiv 1$, $\psi_{(t-1,x)}:= x^e - x^b \equiv 2$, $\psi_{(t-1,y)}:= y^e - y^b \equiv 1$, $\psi_{(t,x)}:= x^e - x^b \equiv 2$, $\psi_{(t,y)}:= y^e - y^b \equiv 2$; $a^e_{t-4}$ represent a evader moving westward, $a^e_{t-3}$ is moving southward, $a^e_{t-2}$ is moving eastward, and $a^e_{t-1}$ is moving northward.
\end{itemize}

\begin{figure*}[t]
        \captionsetup{font=small}
	\centering
	\begin{subfigure}{0.3\linewidth}
		\centering
		\includegraphics[width=\linewidth]{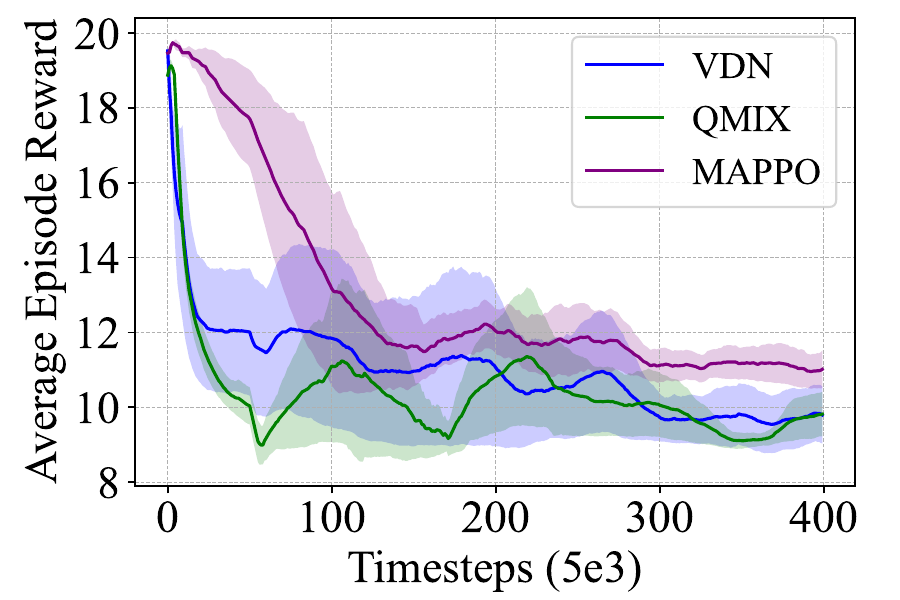}
		\caption{Reward in trigger-embedded 8m}
		\label{chutian3}
	\end{subfigure}
        \vspace{0.5cm}
	\centering
	\begin{subfigure}{0.3\linewidth}
		\centering
		\includegraphics[width=\linewidth]{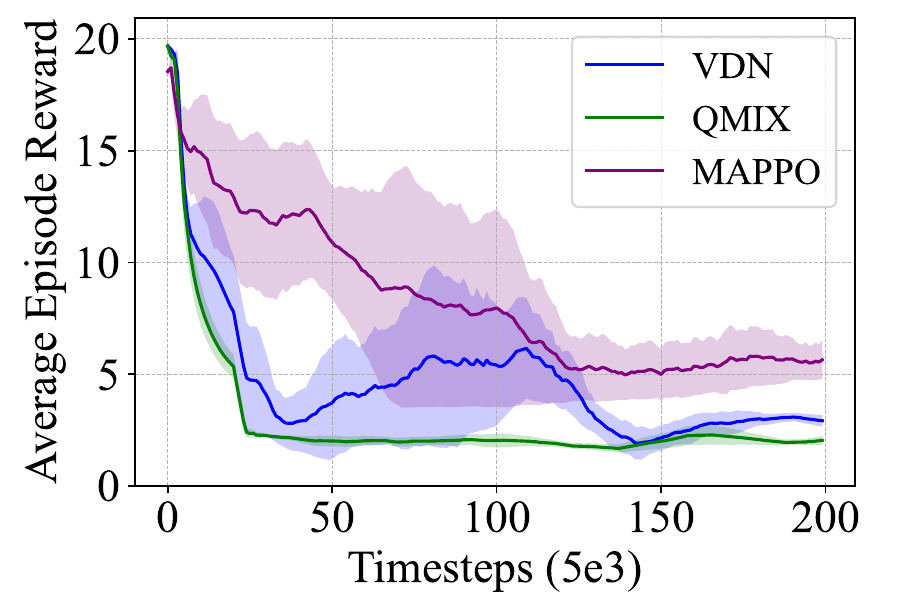}
		\caption{Reward in trigger-embedded 3m}
		\label{chutian3}
	\end{subfigure}
	\centering
	\begin{subfigure}{0.3\linewidth}
		\centering
		\includegraphics[width=\linewidth]{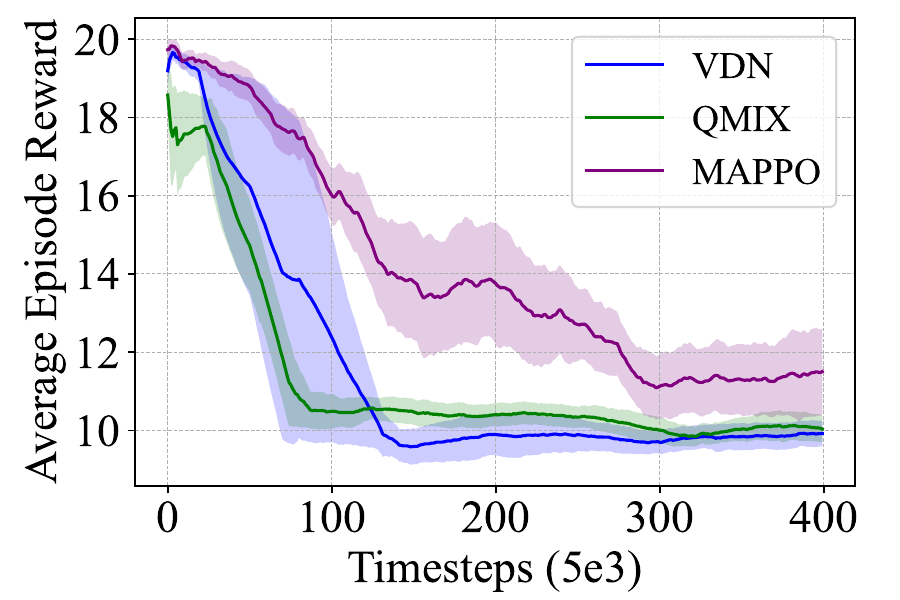}
		\caption{Reward in trigger-embedded 2s3z}
		\label{chutian3}
	\end{subfigure}
        \vspace{0.5cm}
        \centering
	\begin{subfigure}{0.3\linewidth}
		\centering
		\includegraphics[width=\linewidth]{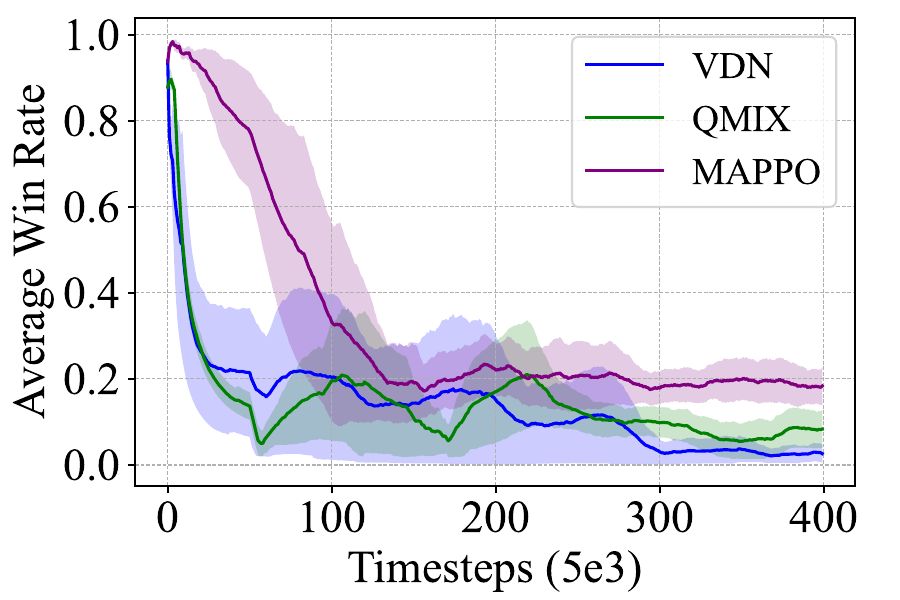}
		\caption{Win rate in trigger-embedded 8m}
		\label{chutian3}
	\end{subfigure}
        \centering
	\begin{subfigure}{0.3\linewidth}
		\centering
		\includegraphics[width=\linewidth]{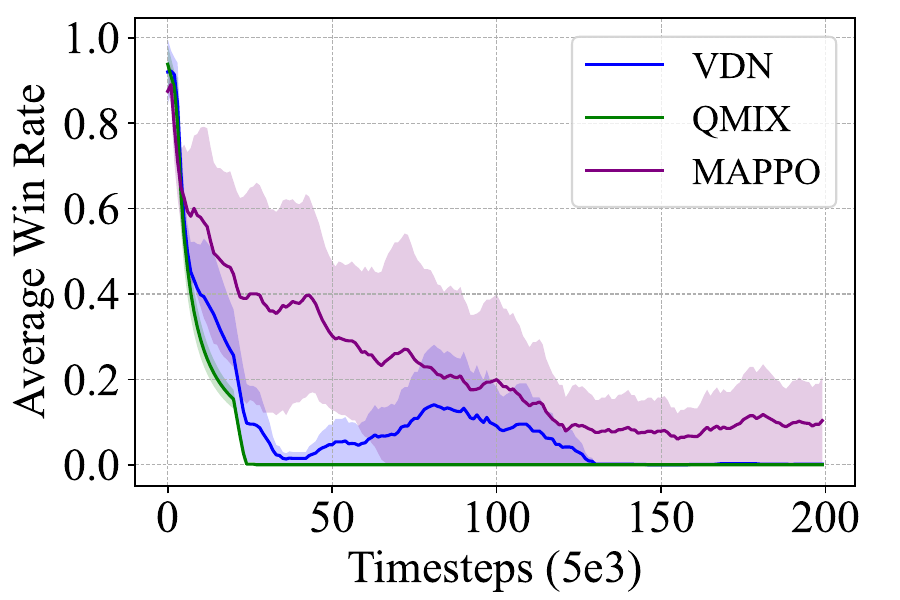}
		\caption{Win rate in trigger-embedded 3m}
		\label{chutian3}
	\end{subfigure}
        \centering
	\begin{subfigure}{0.3\linewidth}
		\centering
		\includegraphics[width=\linewidth]{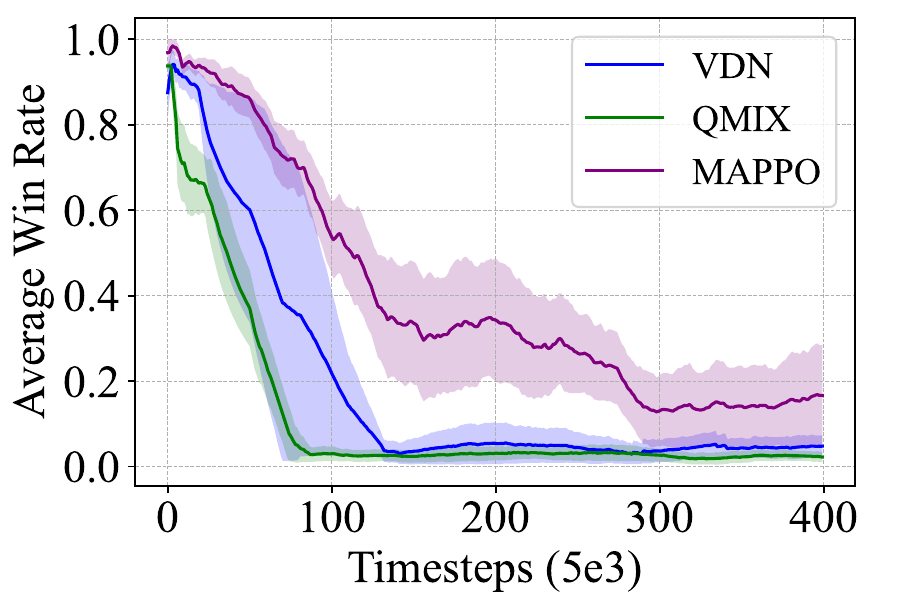}
		\caption{Win rate in trigger-embedded 2s3z}
		\label{chutian3}
	\end{subfigure}
        \vspace{0.5cm}
        \centering
	\begin{subfigure}{0.3\linewidth}
		\centering
		\includegraphics[width=\linewidth]{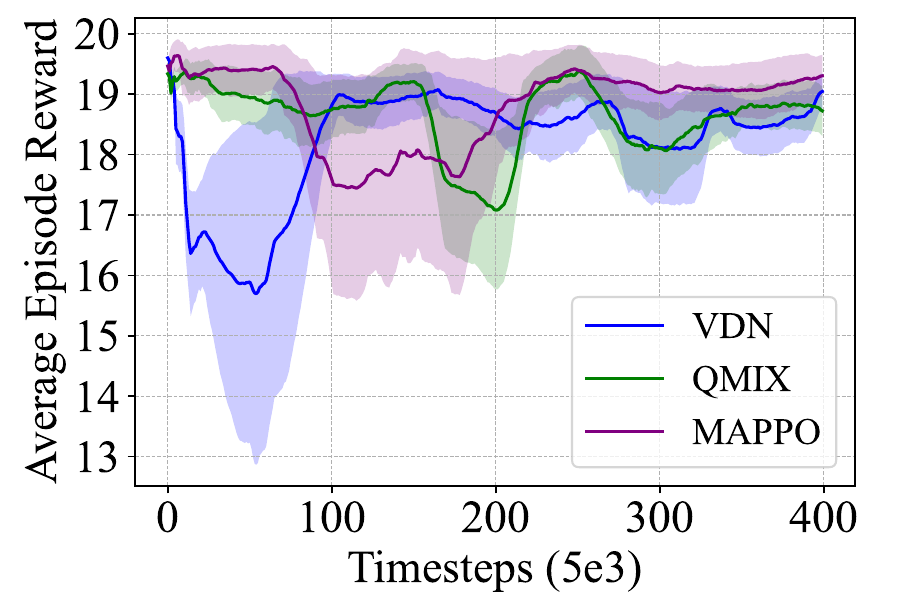}
		\caption{Reward in trigger-free 8m}
		\label{chutian3}
	\end{subfigure}
        \centering
	\begin{subfigure}{0.3\linewidth}
		\centering
		\includegraphics[width=\linewidth]{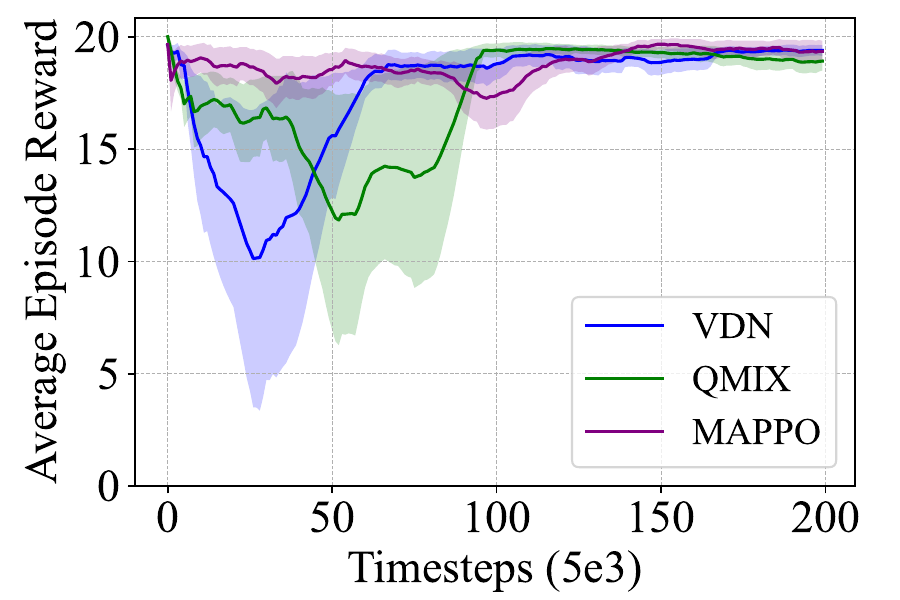}
		\caption{Reward in trigger-free 3m}
		\label{chutian3}
	\end{subfigure}
        \centering
	\begin{subfigure}{0.3\linewidth}
		\centering
		\includegraphics[width=\linewidth]{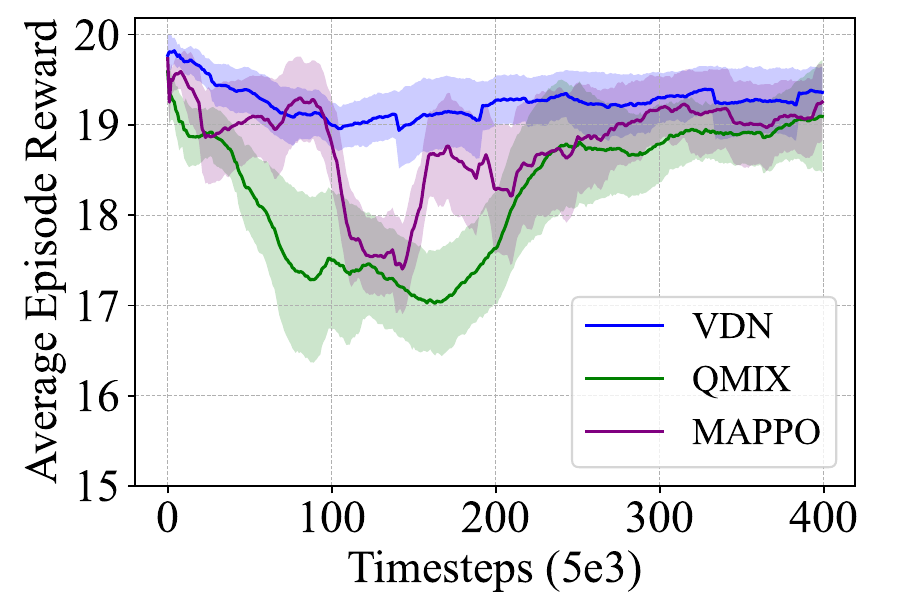}
		\caption{Reward in trigger-free 2s3z}
		\label{chutian3}
	\end{subfigure}
        \vspace{0.5cm}
        \centering
	\begin{subfigure}{0.3\linewidth}
		\centering
		\includegraphics[width=\linewidth]{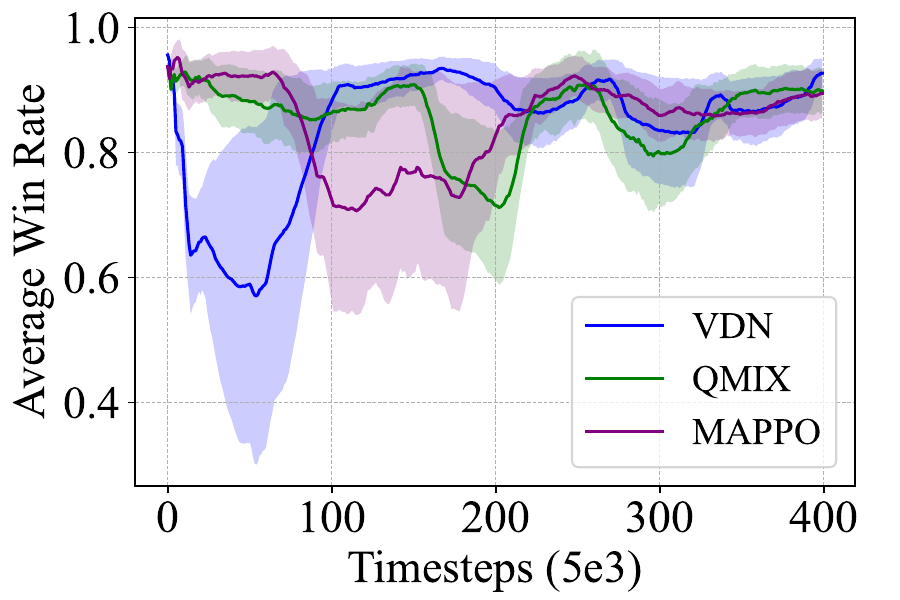}
		\caption{Win rate in trigger-free 8m}
		\label{chutian3}
	\end{subfigure}
        \centering
	\begin{subfigure}{0.3\linewidth}
		\centering
		\includegraphics[width=\linewidth]{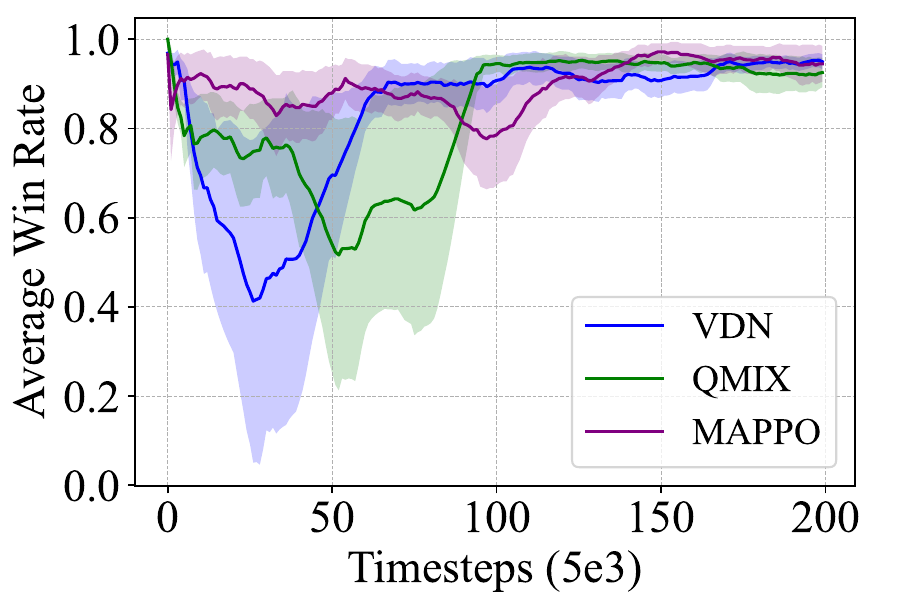}
		\caption{Win rate in trigger-free 3m}
		\label{chutian3}
	\end{subfigure}
        \centering
	\begin{subfigure}{0.3\linewidth}
		\centering
		\includegraphics[width=\linewidth]{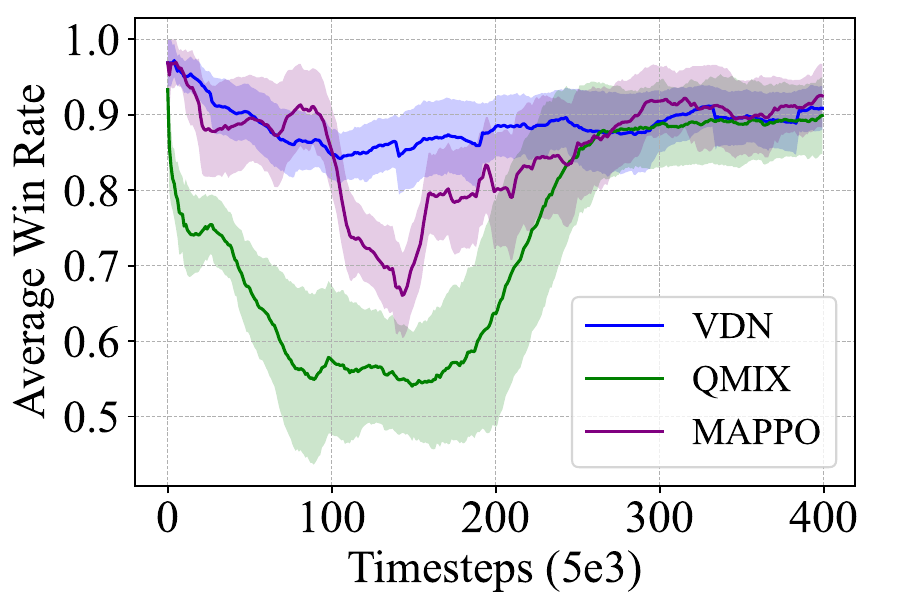}
		\caption{Win rate in trigger-free 2s3z}
		\label{chutian3}
	\end{subfigure}
	\caption{Training curves of average episode rewards and average winning rates of BLAST models attacking against VDN, QMIX, and MAPPO algorithms.}
	\label{SMAC_training}
 \vspace*{-5mm}
\end{figure*}

In each poisoned episode, we search if there is an observable enemy unit (or evader) whose position relative to the unit controlled by the BLAST agent satisfies the first spatial constraint in $\Psi$. If found, we will take control of the enemy unit (or evader) to behave following $\zeta$ defined in the corresponding $\mathcal{T}$, and return it to its heuristic controller after the trigger.
During BLAST model training, we set $\lambda=0.5$ in hacked reward, the size of both clean replay buffer $\mathcal{B}_{c}$ and poisoned replay buffer $\mathcal{B}_{p}$ to 5000, the size of batch to 32, the discount factor $\gamma=0.99$, and the learning rate $\alpha=5e^{-4}$. For VDN and QMIX, we set the update frequency of the target network to 200 steps, and the greedy factor $\sigma=0.05$. For MAPPO, we set the GAE parameter $\epsilon=0.2$. For the 8m and 3m maps, we set the poisoning rate $p=0.05$, and the attack period $L = 20$. For 2s3z map, we set $p=0.02$ and $L = 40$. For the Pursuit environment, we set $p=0.05$ and $L = 40$.

\subsubsection{Evaluation Metrics} We use the metrics in Table \ref{metrics} to evaluate the effectiveness and stealthiness of BLAST. These metrics quantify the impact of BLAST on DRL policies, where the higher ASR and AER values are, the more effective BLAST is, and the lower CPVR and WRVR values are, the more stealthy BLAST is.

\subsection{Attack Results in SMAC}
\label{subsec:smac}

\begin{table*}[thbp]
\captionsetup{font=small}
\centering
    \caption{Attacks performance of BLAST attack against VDN, QMIX, and MAPPO algorithms. }
    \label{tab:performance}
    \renewcommand{\arraystretch}{1.5}
    \setlength{\tabcolsep}{7pt}
    \begin{tabular}{cccccccccccc}
    \toprule 
    Map                   & Algorithm & $cER$$\uparrow$ & $cWR$$\uparrow$  & $bER$$\uparrow$ & $bWR$$\uparrow$  & ${bER}_{tg}$$\downarrow$ & ${bWR}_{tg}$$\downarrow$ & $AER$$\uparrow$  & $ASR$$\uparrow$   & $CPVR$$\downarrow$ & $WRVR$$\downarrow$ \\ \hline
    \multirow{3}{*}{8m}   & VDN       &19.55&95.6\%&19.23&93.7\%& 10.05  & 3.1\%  &48.6\%&96.7\% &1.6\% &2.0\% \\
                          & QMIX      &19.73&96.8\%&19.01&93.5\%& 9.67   & 4.6\%  &51.0\%&95.1\% &3.7\% &3.4\% \\
                          & MAPPO     &19.81&97.5\%&19.36&91.6\%& 11.24  & 18.4\% &43.3\%&81.1\% &2.3\% &6.1\% \\ \hline
    \multirow{3}{*}{3m}   & VDN       &19.64&97.4\%&19.35&95.5\%& 2.73   &   0    &86.1\%&100.0\%&1.5\% &1.9\%       \\
                          & QMIX      &19.81&98.3\%&19.21&95.2\%& 2.28   &   0    &88.5\%&100.0\%&3.0\% &3.2\% \\
                          & MAPPO     &19.68&97.5\%&19.30&94.1\%& 5.92   & 13.2\% &69.9\%&86.5\% &2.0\% &3.5\% \\ \hline
    \multirow{3}{*}{2s3z} & VDN       &19.72&95.1\%&19.39&93.6\%& 9.98   & 4.3\%  &49.4\%&95.5\% &1.7\% &4.6\% \\
                          & QMIX      &19.80&95.9\%&19.04&92.8\%& 9.86   & 1.9\%  &50.2\%&98.0\% &3.8\% &3.2\% \\
                          & MAPPO     &19.93&97.9\%&19.37&90.8\%& 11.70  & 16.7\% &41.3\%&83.0\% &2.8\% &7.3\% \\
    \bottomrule 
    \end{tabular}
\end{table*}

Fig. \ref{SMAC_training} illustrates how the average episode rewards and the average winning rate change during the training of the BLAST models attacking VDN, QMIX, and MAPPO. In Fig. \ref{SMAC_training}(a)-(f), we can see that the average episode rewards and the winning rates in poisoned episodes which are trigger-embedded decrease with the number of training steps. Affected by backdoor learning, the episode rewards and the winning rates in clean episodes which are trigger-free first decrease, and as learning progresses, they will rise again to levels similar to before training, as shown in Fig. \ref{SMAC_training}(g)-(l). After the BLAST model training process is completed, more specific evaluation metrics are shown in Table \ref{tab:performance}. We can see that metrics ${bER}_{tg}$ and ${bWR}_{tg}$ are significantly lower compared to metrics $cER$ and $cWR$, while metrics $bER$ and $bWR$ are only slightly lower compared to metrics $cER$ and $cWR$. In other words, metrics $AER$ and $ASR$ are very high while metrics $CPVR$ and $WRVR$ are maintained at a very low level. In the attacks against multiple algorithms in multiple maps, the $AER$ and $ASR$ can be as high as 88.5\% and 100.0\%, respectively, while the $CPVR$ and $WRVR$ can be as low as 1.5\% and 1.9\%, respectively, which proves the effectiveness, stealthiness, and universality of our backdoor attacks. In addition, we can see that the performance of the attacks against VDN and QMIX is better than that against MAPPO, which indicates that MAPPO is a bit more robust against our attacks.

\begin{figure}[h!]
    \captionsetup{font=small}
    \centering
    \centering
    \begin{subfigure}{.48\textwidth}
        \centering
        \includegraphics[width=\linewidth]{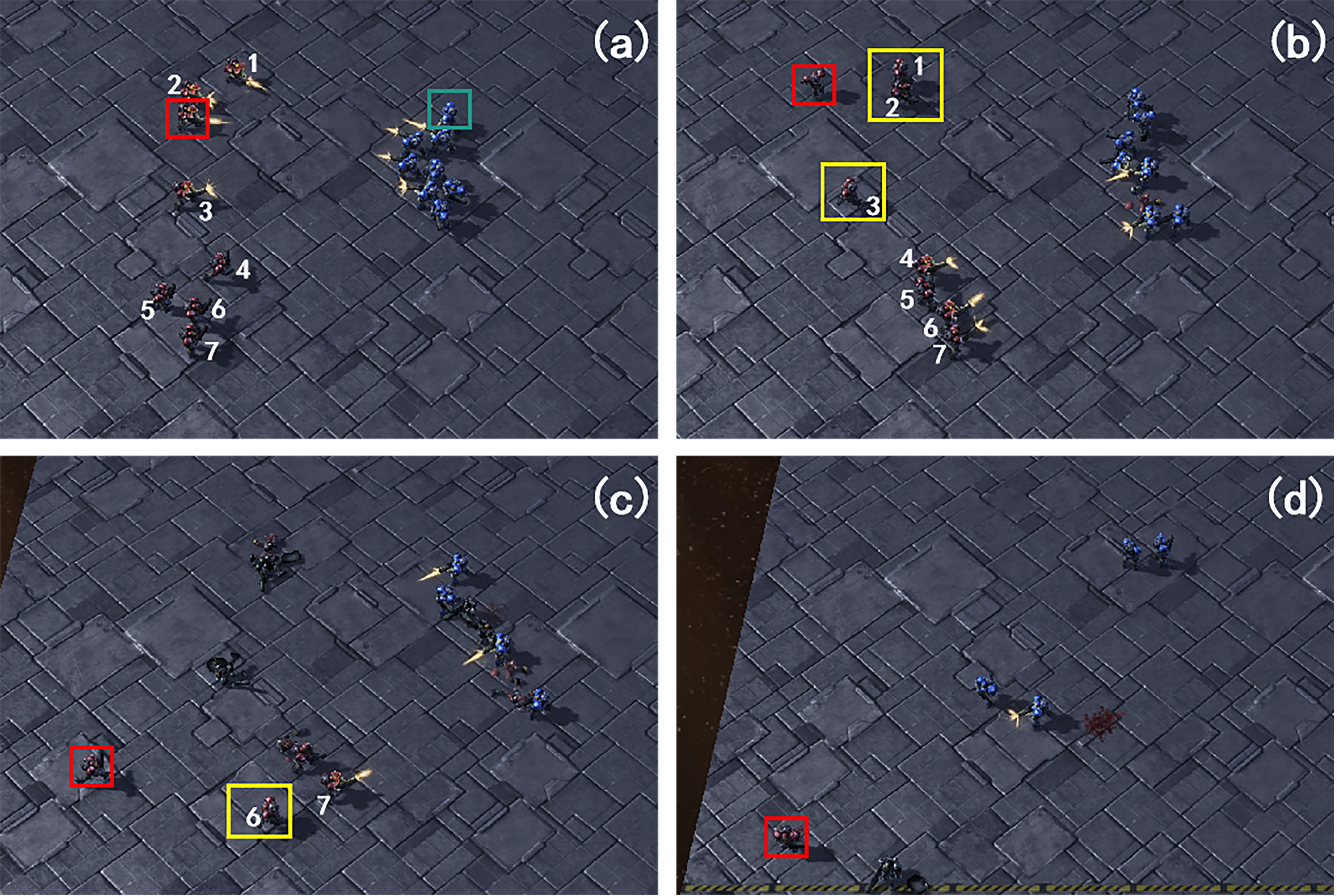}
        \caption{8m map}
    \end{subfigure}%

    \medskip 
    \vspace{0.18cm}

    \begin{subfigure}{.48\textwidth}
        \centering
        \includegraphics[width=\linewidth]{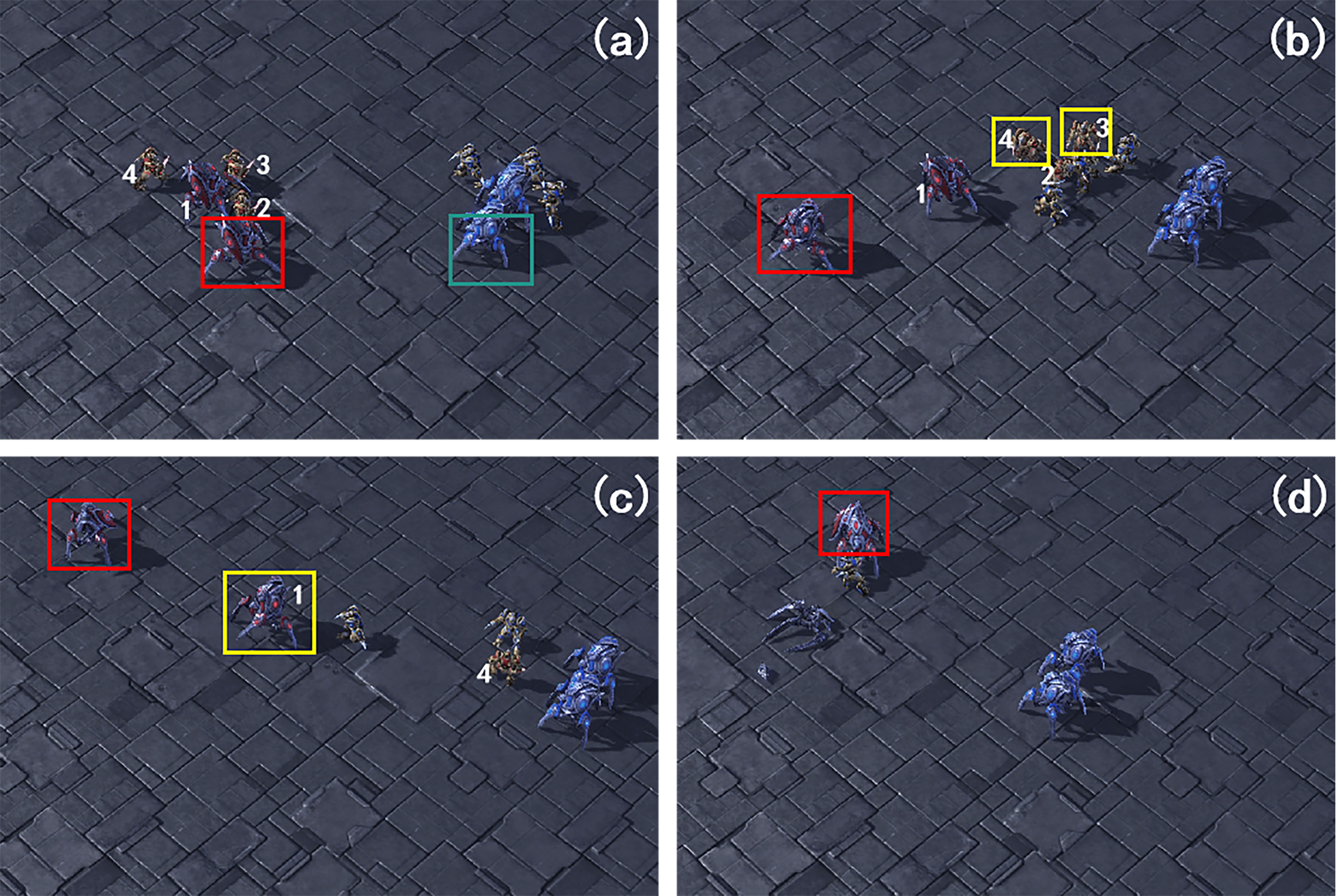}
        \caption{2s3z map}
    \end{subfigure}
    \caption{The snapshots of agents' behaviors in a poisoned episode. The green, red, and yellow boxes denote the trigger, the BLAST agent, and the affected clean agents. Numbers represent the ID of clean agents.}
    \label{map_behaviors}
    \vspace{2mm}
\end{figure}

\begin{figure}[h!]
    \captionsetup{font=small}
    \centering
    \begin{subfigure}{.48\textwidth}
        \centering
        \includegraphics[width=\linewidth]{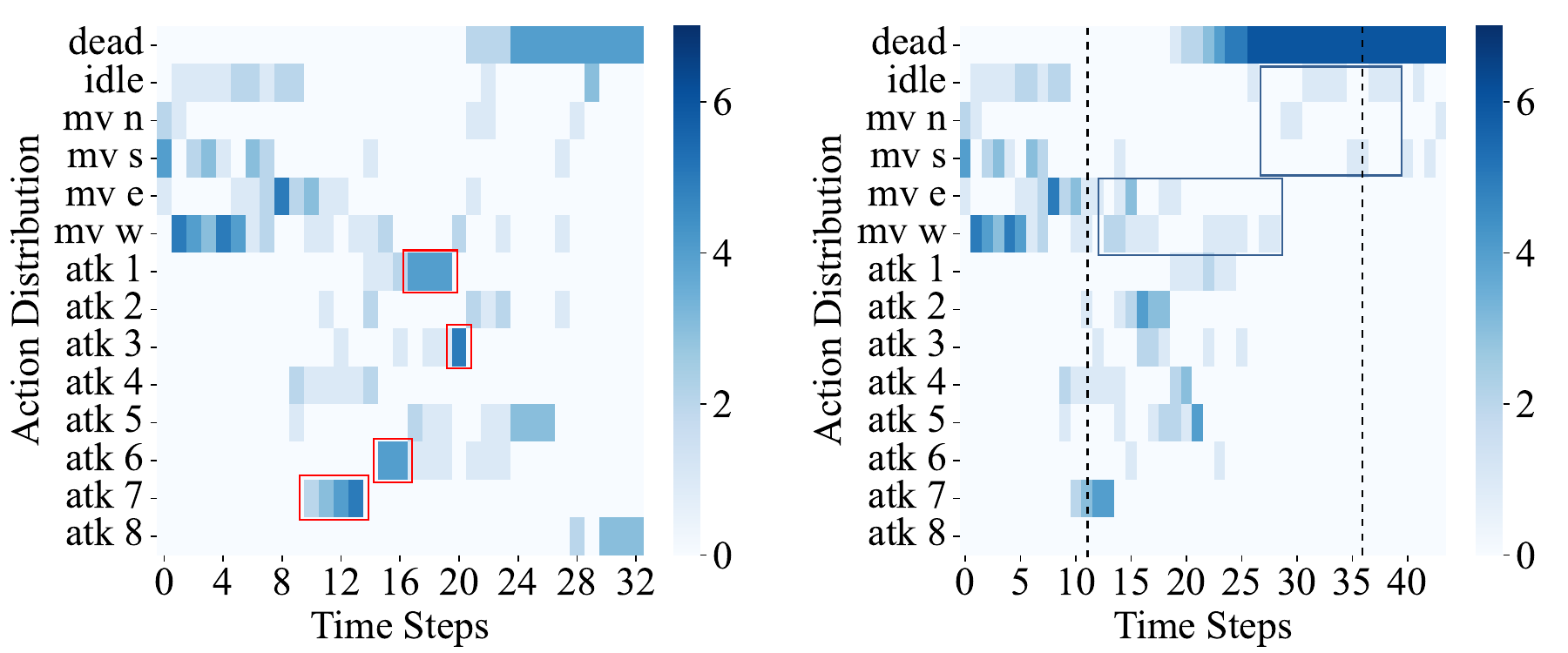}
        \caption{Clean (left) and poisoned (right) episode in 8m map}
    \end{subfigure}%

    \medskip
    \vspace{0.1cm}

    \begin{subfigure}{.48\textwidth}
        \centering
        \includegraphics[width=\linewidth]{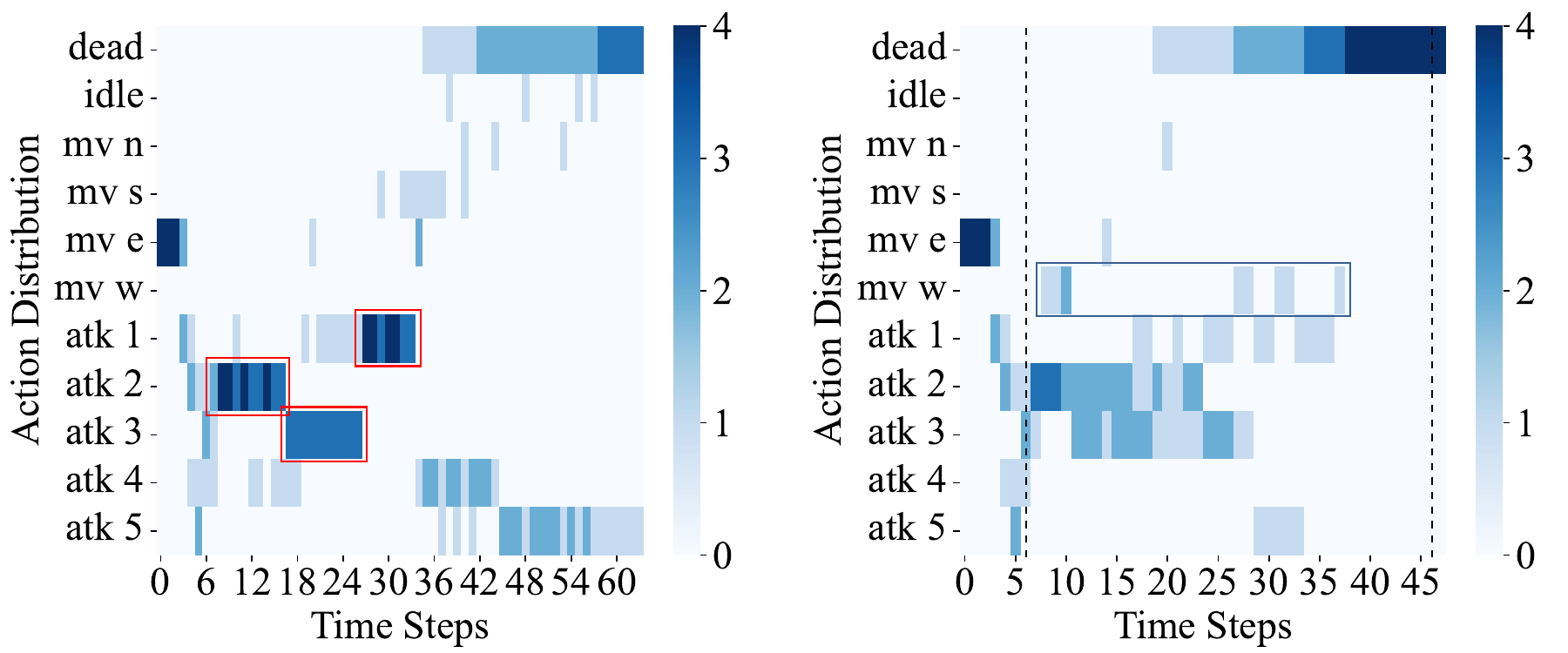}
        \caption{Clean (left) and poisoned (right) episode in 2s3z map}
    \end{subfigure}
    \caption{Action distribution of all clean agents in a clean episode and a poisoned episode. The red and blue boxes represent the behavior of concentrated fire and an increase in moving behaviors, respectively. The time steps between the two black dotted lines indicate the attack period.}
    \label{actions_hot}
\end{figure}

We further analyze the behaviors of the BLAST agent and clean agents during the attack period to explore how the BLAST agent affects the clean agents and thus leads to the team's failure. Fig. \ref{map_behaviors} shows the snapshots of episodes of VDN agents in 8m and 2s3z maps, respectively.
Firstly, we control an enemy unit to execute the trigger behavior. Once the trigger behavior is fully executed, the backdoor in the BLAST agent is triggered, as shown in Fig. \ref{map_behaviors}(a)-(a) and \ref{map_behaviors}(b)-(a).
Then, the BLAST agent begins to perform malicious actions, specifically leaving the main battlefield and moving within the sight of its teammates, successfully inducing some teammates to move to unfavorable positions, resulting in other teammates being attacked by enemy units, as shown in Fig. \ref{map_behaviors}(a)-(b-c) and \ref{map_behaviors}(b)-(b-c).
In the end, all teammates of the BLAST agent die, followed by the death of the BLAST agent itself, and the agent team fails, as shown in Fig. \ref{map_behaviors}(a)-(d) and \ref{map_behaviors}(b)-(d).

Meanwhile, we analyze the difference in the joint actions distribution of all clean agents in a clean and a poisoned episode, as shown in Fig. \ref{actions_hot}.
We can notice that in the clean episode, the agents show concentrated fire behavior, quickly killing the enemy units one by one. In the end, there are still multiple agents surviving.
In the poisoned episode, before the attack begins, the clean agents' action distribution is the same as in the clean episode. During the attack period, the clean agents are affected by the BLAST agent and begin to exhibit abnormal behavior, including no longer performing concentrated fire, more dispersed action distribution, reduced attack actions, and increased movement actions (as shown in the blue boxes in Fig. \ref{actions_hot}). These changes lead to the deaths of all agents and the agent team's failure.

\subsection{Attack Results in Pursuit}
\label{subsec:pursuit}
\begin{figure}[!t]
        \captionsetup{font=small}
	\centering
	\begin{subfigure}{0.48\linewidth}
		\centering
            \includegraphics[width=\linewidth]{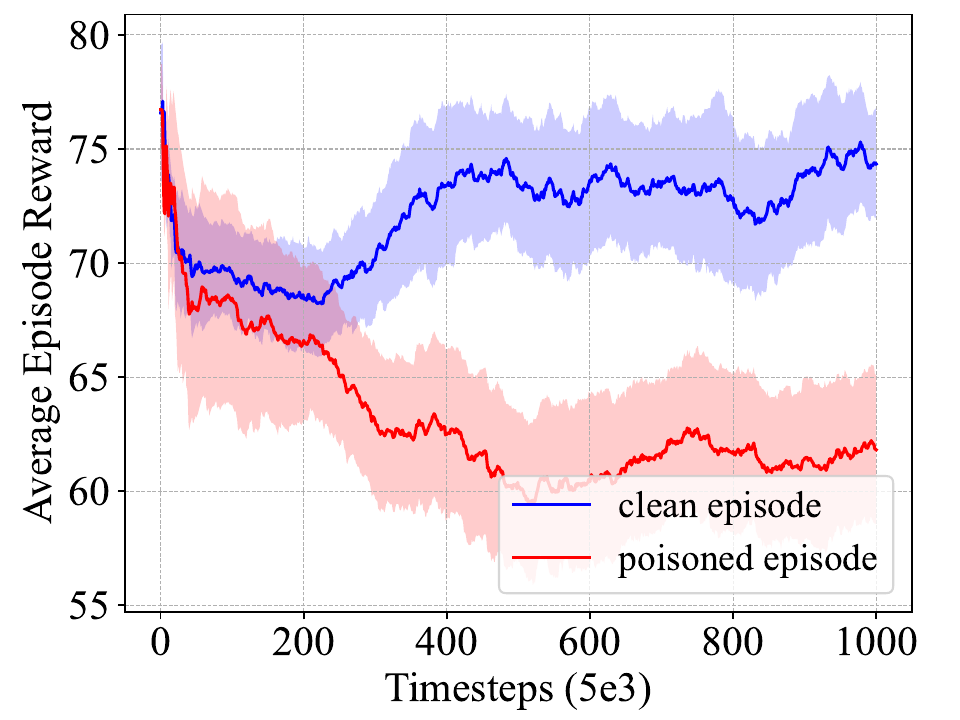}
		\caption{VDN}
	\end{subfigure}
	\centering
	\begin{subfigure}{0.48\linewidth}
		\centering
		\includegraphics[width=\linewidth]{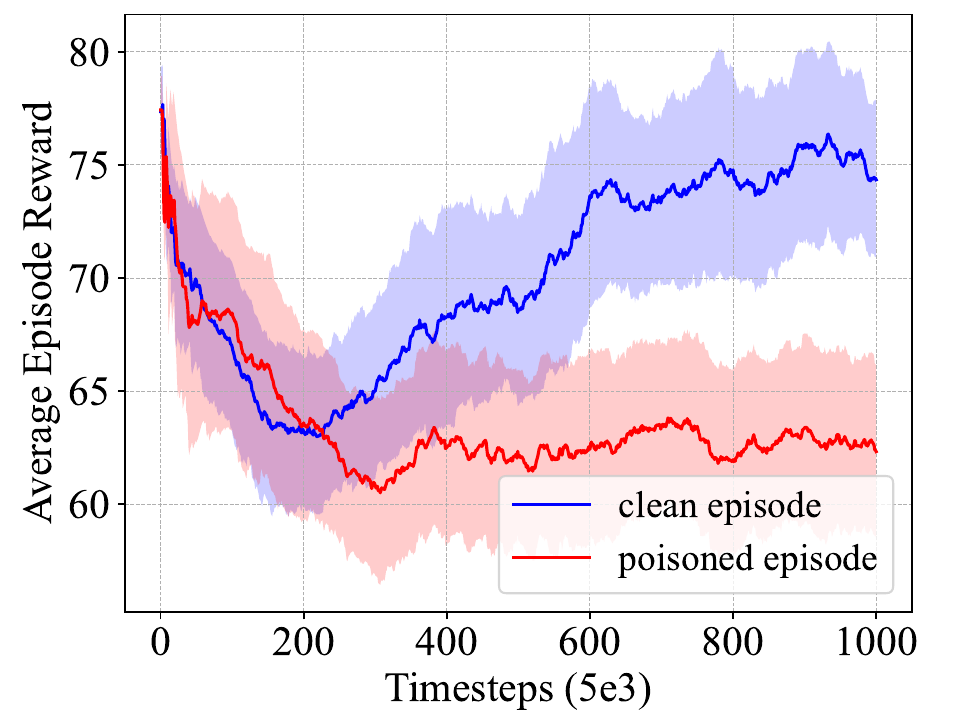}
		\caption{QMIX}
	\end{subfigure}
	\caption{The average episode rewards of the BLAST model of VDN and QMIX.}
	\label{pursuit_training}
\end{figure}

\begin{figure}[!t]
    \captionsetup{font=small}
    \centering
    \begin{subfigure}{.48\textwidth}
        \centering
        \includegraphics[width=\linewidth]{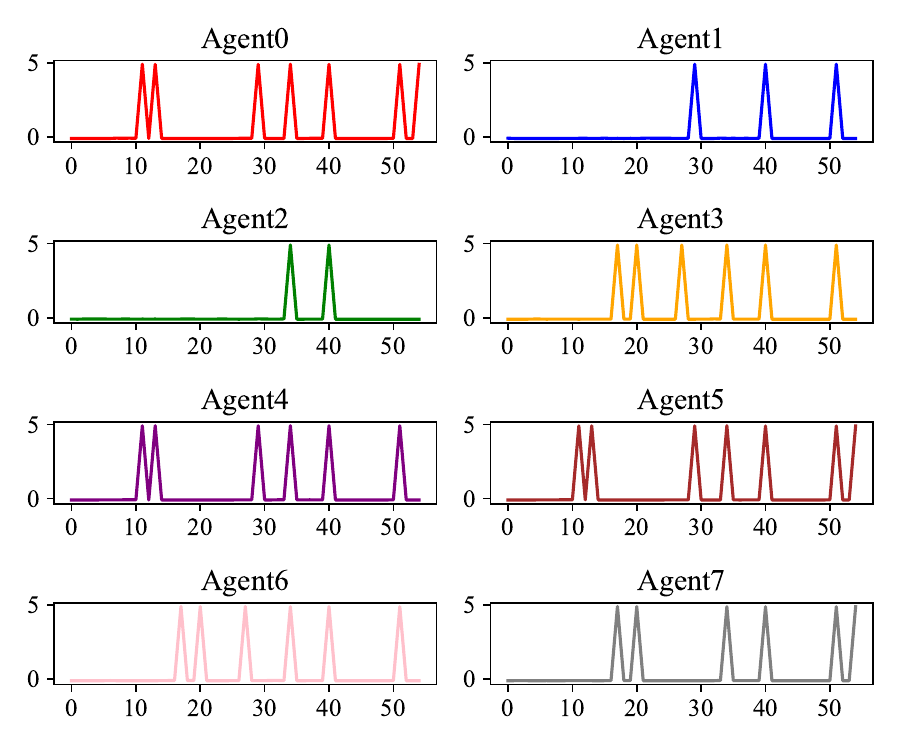}
        \caption{In a clean episode}
    \end{subfigure}

    \vspace{2mm}
    \begin{subfigure}{.48\textwidth}
        \centering
        \includegraphics[width=\linewidth]{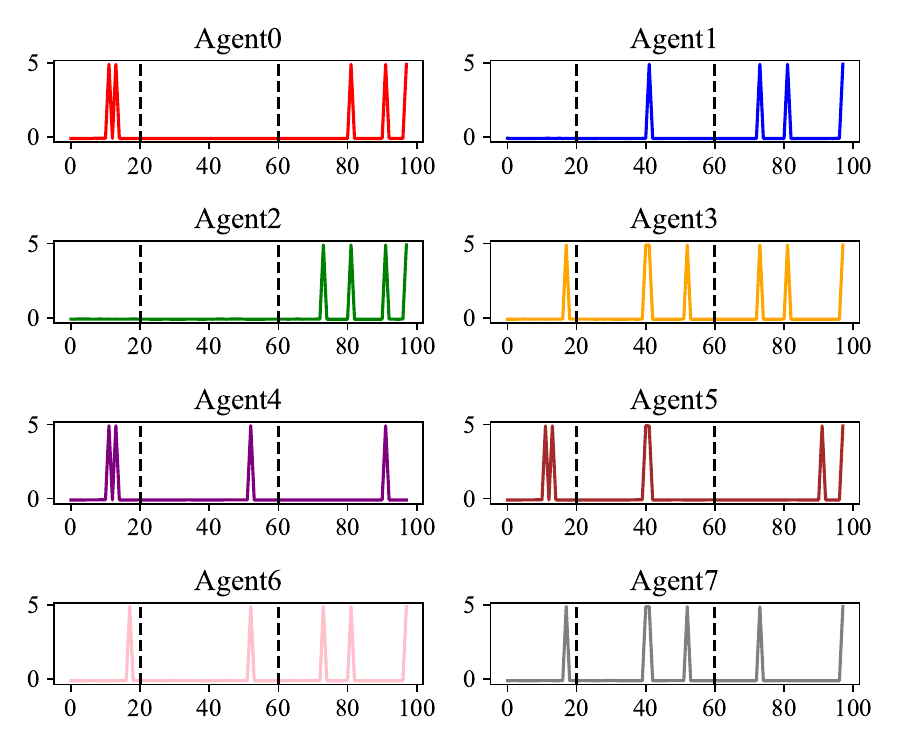}
        \caption{In a poisoned episode}
    \end{subfigure}
    \caption{Analysis of the time-step reward of every agent. The time steps between the two black dotted lines indicate the attack period. Where Agent0 is the BLAST agent.}
    \label{pursuit-time-step-reward}
\end{figure}

To verify the universality of our BLAST attacks, we conduct BLAST in the Pursuit environment \cite{pursuit}. Fig. \ref{pursuit_training} shows the training process of BLAST policies of VDN and QMIX. We can see that the average episode reward in poisoned episodes gradually decreases as training progresses, while the average episode reward in clean episodes eventually converges to the original level. This indicates that our BLAST backdoor attacks are effective.

We further analyze the time-step reward of each agent in clean and poisoned episodes under the same random seed in Fig. \ref{pursuit-time-step-reward}. Note that the Pursuit environment supports viewing the individual reward value of each agent and the team reward is the average of the reward values of each agent.
An increase in the reward of a certain agent indicates that it participates in capturing an evader and successfully captures it.
It is not difficult to observe that the rewards of clean agents are not affected before the BLAST attack begins. In the poisoned episode, the reward values of all clean agents during the attack period decrease compared to the clean episode. This indicates that during the attack period, these agents do not collaborate well to capture evaders; that is, the BLAST agent plays a unilateral malicious role. After the attack period, we can find that most agents do not quickly experience an increase in rewards. This is because due to the lag effect of the BLAST attack (\textit{i.e.}, the BLAST agent guides some of the clean agents to the unfavorable position), the agents cannot immediately capture the remaining evaders, but need some time to do so. Once the BLAST attack is performed successfully, the agent team will spend more time capturing all the evaders.

\subsection{Resistance to Backdoor Defense Methods}

\label{subsec:defense}
\begin{figure}[!t]
        \captionsetup{font=small}
	\centering
	\begin{subfigure}{0.48\linewidth}
		\centering
		\includegraphics[width=\linewidth]{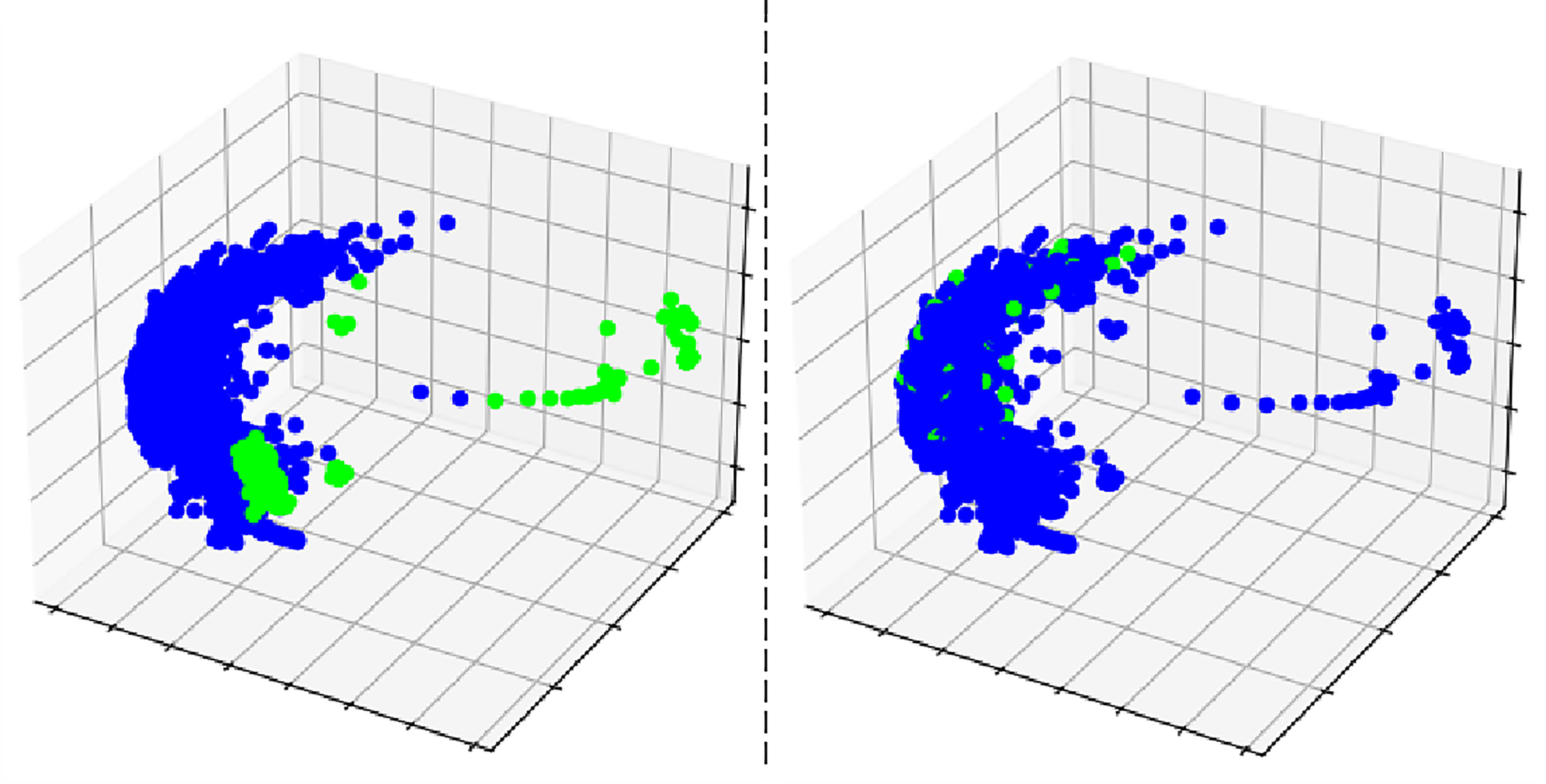}
		\caption{IDLE (Pre vs. GT)}
		\label{chutian3}
	\end{subfigure}
        \vspace{0.18cm}
	\centering
	\begin{subfigure}{0.48\linewidth}
		\centering
		\includegraphics[width=\linewidth]{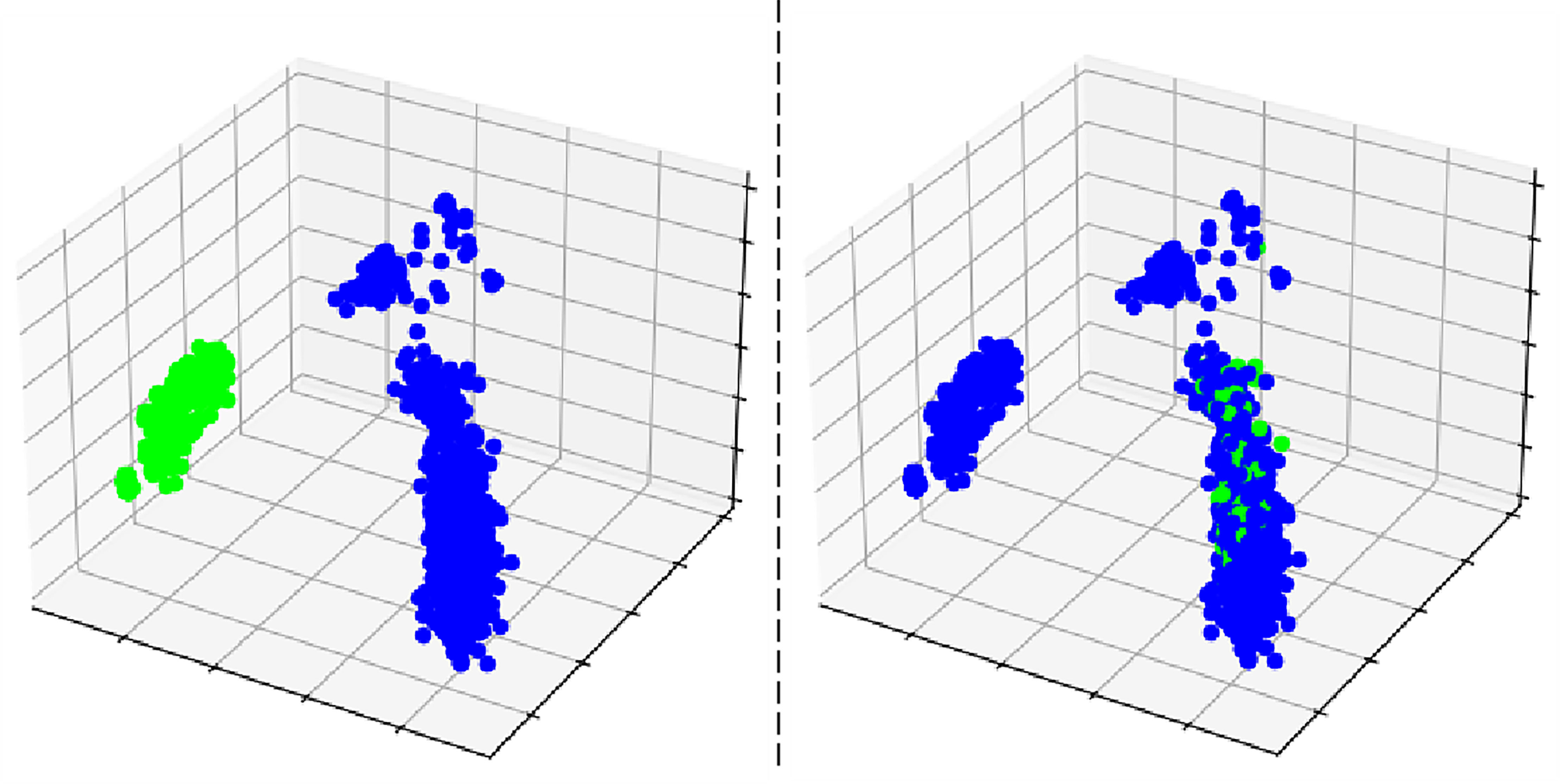}
		\caption{Move northward (Pre vs. GT)}
		\label{chutian3}
	\end{subfigure}
        \vspace{0.18cm}
	\centering
	\begin{subfigure}{0.48\linewidth}
		\centering
		\includegraphics[width=\linewidth]{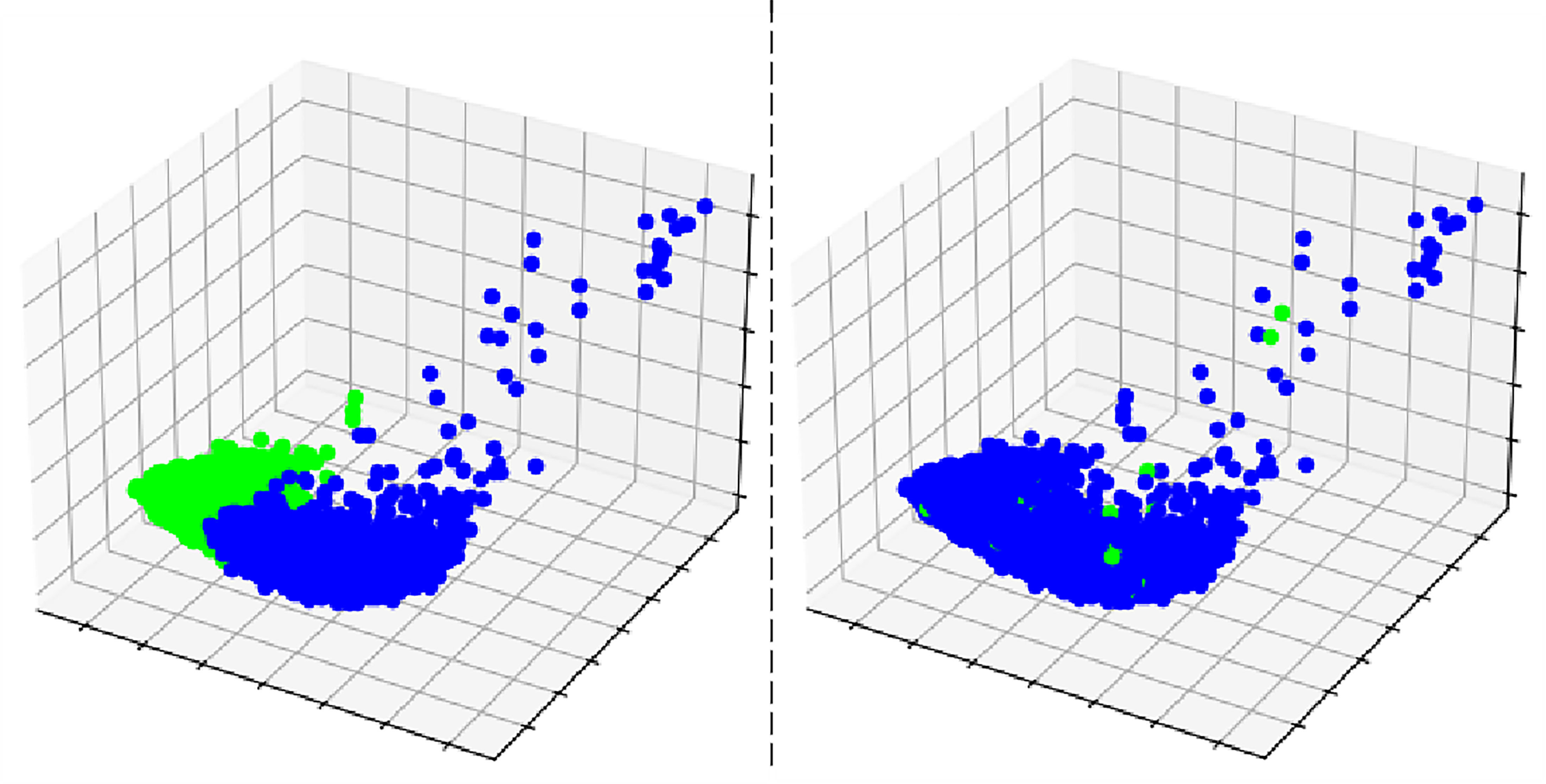}
		\caption{Move southward (Pre vs. GT)}
		\label{chutian3}
	\end{subfigure}
        \centering
	\begin{subfigure}{0.48\linewidth}
		\centering
		\includegraphics[width=\linewidth]{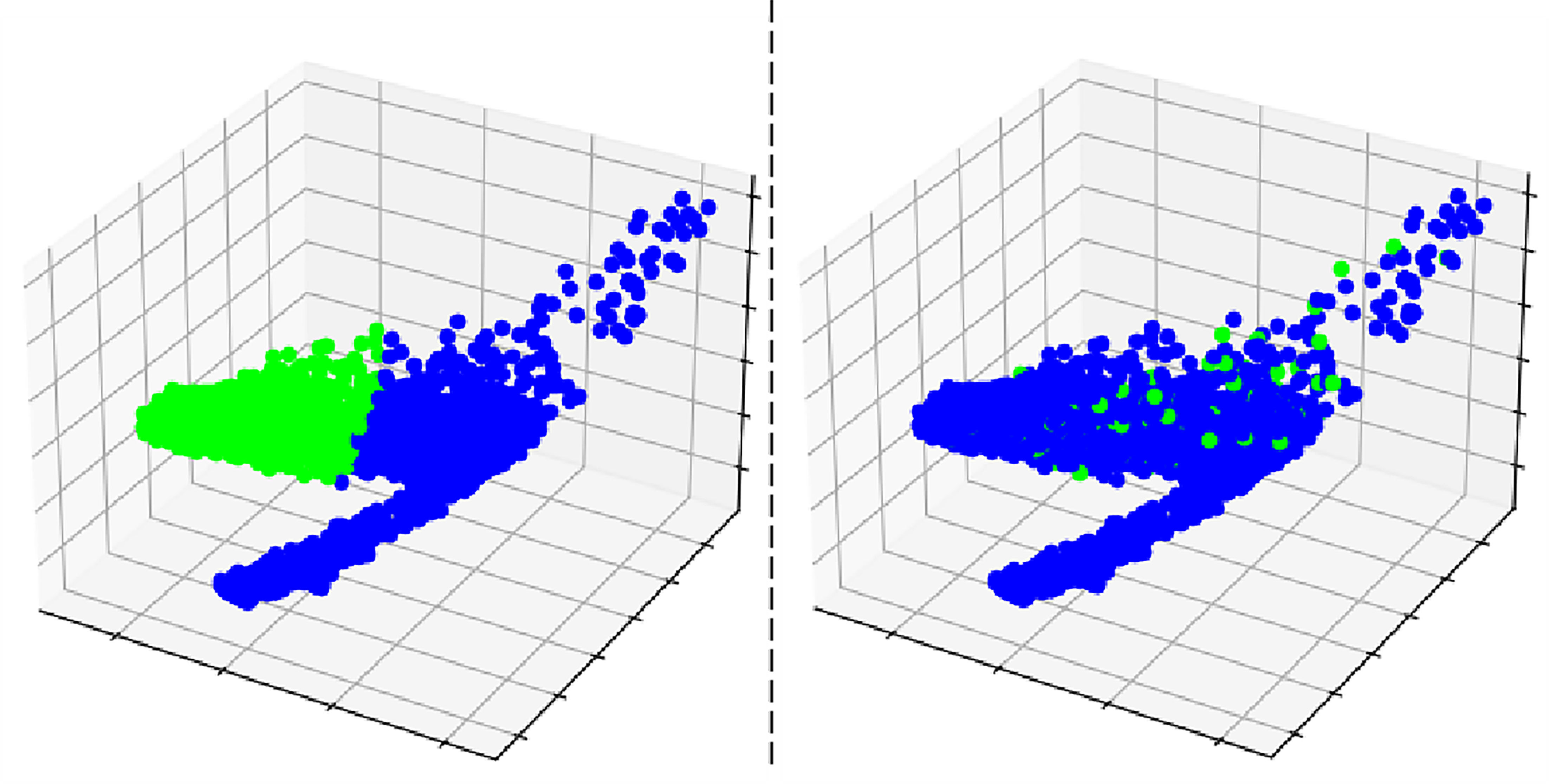}
		\caption{Move eastward (Pre vs. GT)}
		\label{chutian3}
	\end{subfigure}
        \vspace{0.18cm}
        \centering
	\begin{subfigure}{0.48\linewidth}
		\centering
		\includegraphics[width=\linewidth]{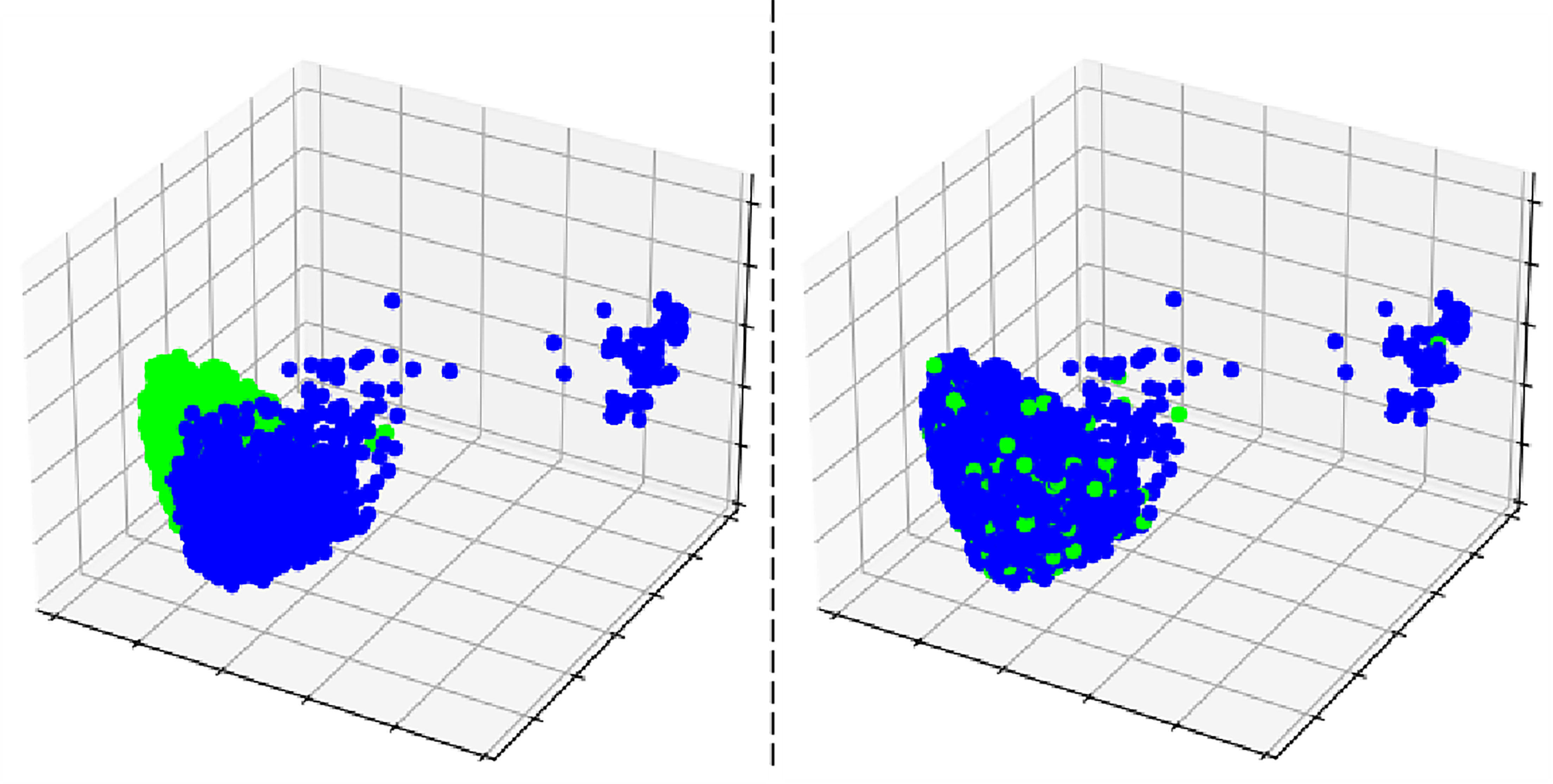}
		\caption{Move westward (Pre vs. GT)}
		\label{chutian3}
	\end{subfigure}
        \centering
	\begin{subfigure}{0.48\linewidth}
		\centering
		\includegraphics[width=\linewidth]{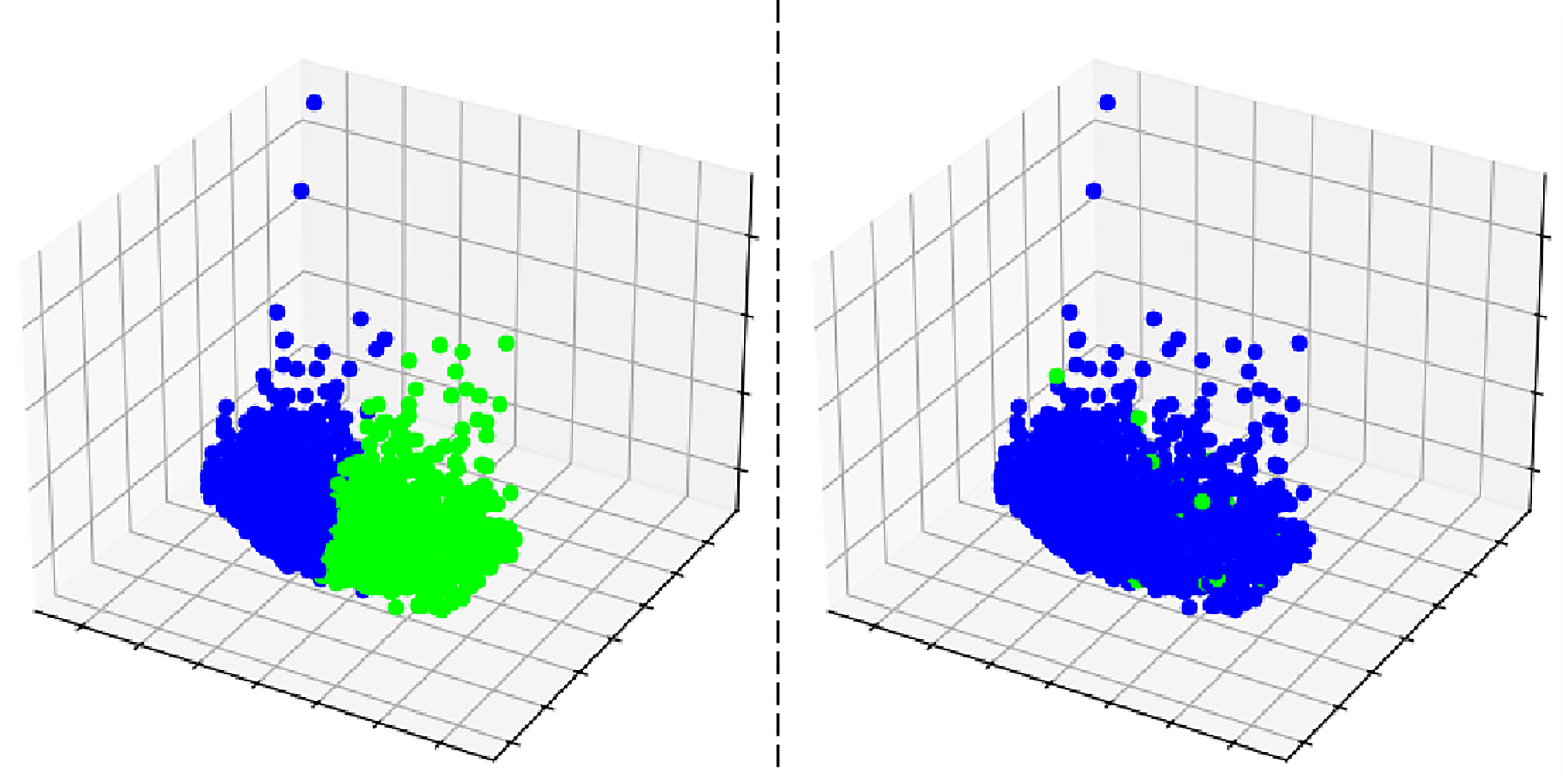}
		\caption{Attack enemy 2 (Pre vs. GT)}
		\label{chutian3}
	\end{subfigure}
	\caption{Backdoor detection results via activation clustering \cite{chen2019ac}. ``Pre'' is the prediction of activation clustering. ``GT'' means groundtruth.}
        \label{ac}
        \vspace{-2mm}
\end{figure}

\begin{figure}[!t]
        \captionsetup{font=small}
	\centering
	\begin{subfigure}{0.48\linewidth}
		\centering
		\includegraphics[width=\linewidth]{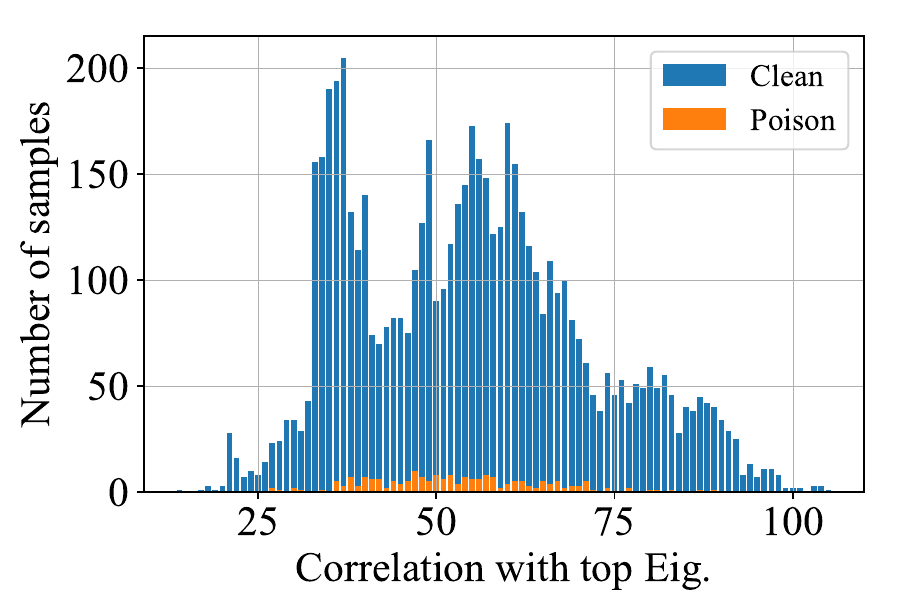}
		\caption{IDLE}
		\label{chutian3}
	\end{subfigure}
        \vspace{0.15cm}
	\centering
	\begin{subfigure}{0.48\linewidth}
		\centering
		\includegraphics[width=\linewidth]{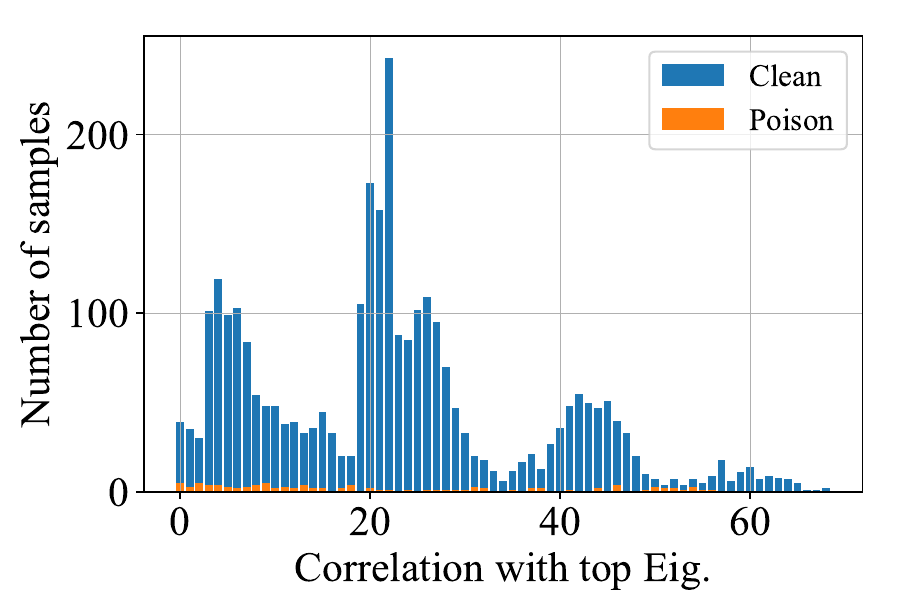}
		\caption{Move northward}
		\label{chutian3}
	\end{subfigure}
        \vspace{0.15cm}
	\centering
	\begin{subfigure}{0.48\linewidth}
		\centering
		\includegraphics[width=\linewidth]{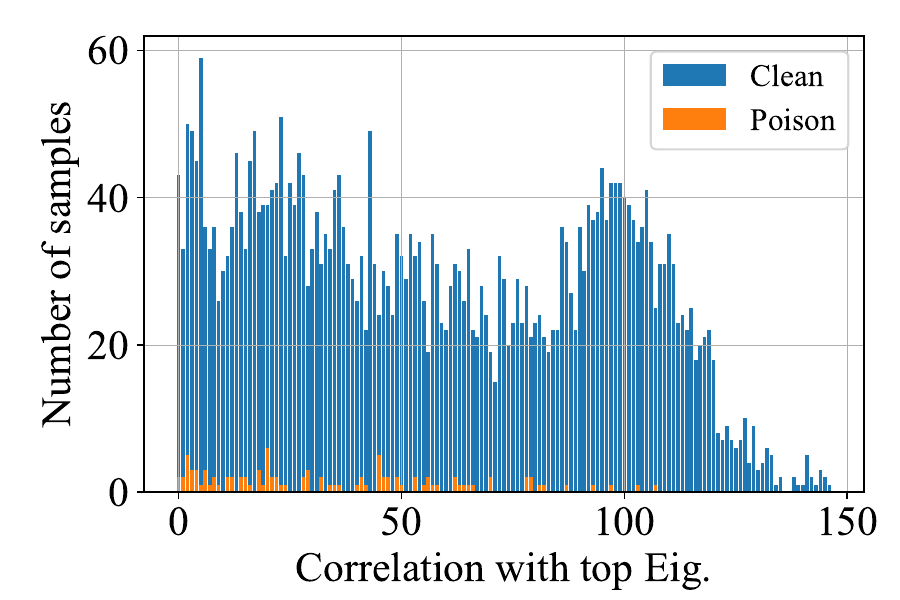}
		\caption{Move southward}
		\label{chutian3}
	\end{subfigure}
        \centering
	\begin{subfigure}{0.48\linewidth}
		\centering
		\includegraphics[width=\linewidth]{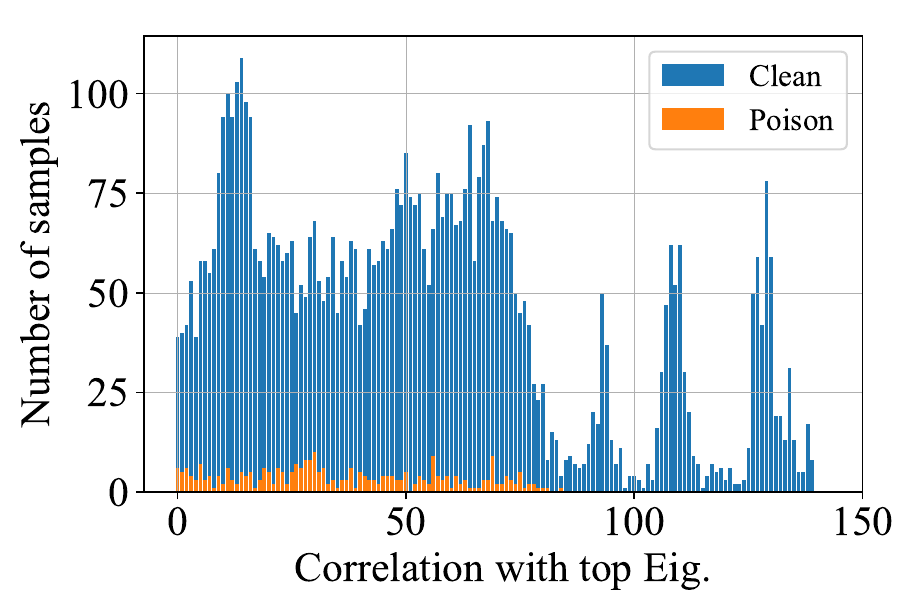}
		\caption{Move eastward}
		\label{chutian3}
	\end{subfigure}
        \vspace{0.15cm}
        \centering
	\begin{subfigure}{0.48\linewidth}
		\centering
		\includegraphics[width=\linewidth]{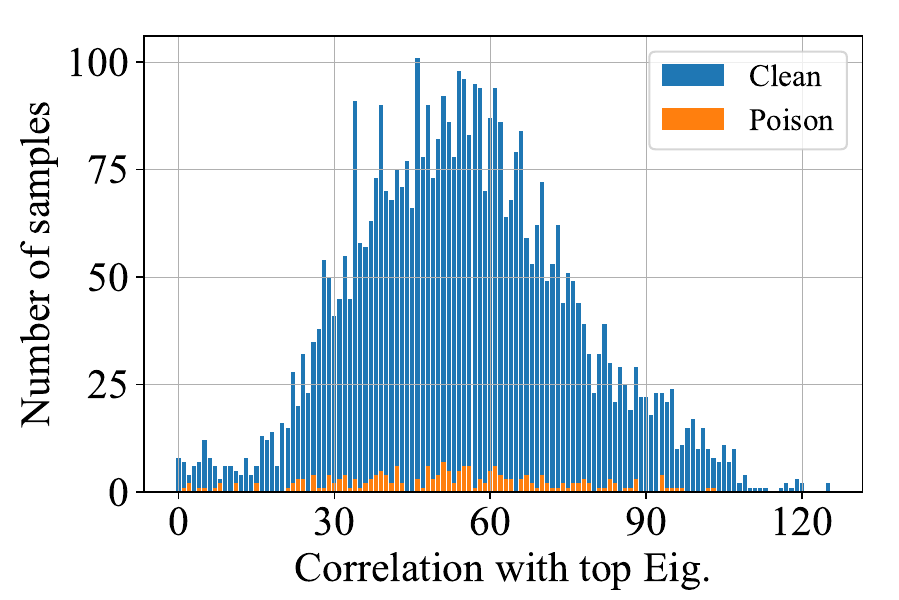}
		\caption{Move westward}
		\label{chutian3}
	\end{subfigure}
        \centering
	\begin{subfigure}{0.48\linewidth}
		\centering
		\includegraphics[width=\linewidth]{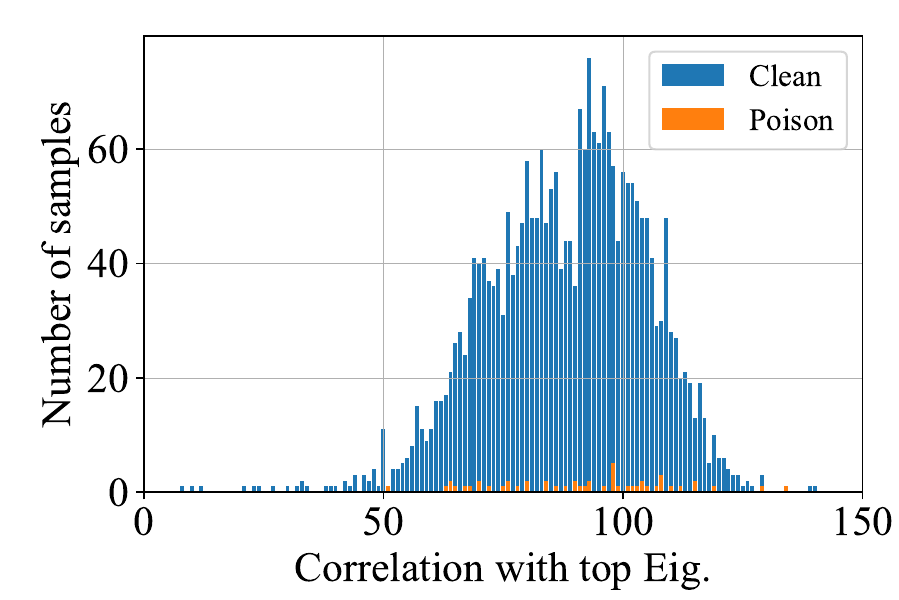}
		\caption{Attack enemy 2}
		\label{chutian3}
	\end{subfigure}
	\caption{Backdoor detection results via spectral signature \cite{SpectralSignatures}.}
	\label{ss}
\end{figure}

To validate the stealthiness of our BLAST attack, we investigate its detectability against existing defense mechanisms in VDN. Three established approaches are evaluated: activation clustering \cite{chen2019ac}, spectral signature analysis \cite{SpectralSignatures} (both designed for DNNs), and BIRD \cite{DBLP:conf/nips/Chen000S23} (developed for DRL). We use the BLAST policy to interact with the 8m environment and sample 28000 clean samples and 950 poisonous samples according to action types, dividing them into six sets (including 5 movement actions and 1 attack action). Following activation clustering, we extract the activations in the penultimate layer of the model and use principal component analysis to downscale the activations to 3 dimensions and then cluster them using k-means with $k = 2$. The clustering analysis results are shown in Fig. \ref{ac}, in which we can discover that the activations of poisonous samples are not distinguishable from the clean ones, and the prediction results are incorrect. Besides, the detection results of spectral signature are shown in Figure 9, where we can find that the distribution of outlier scores for the poisonous samples is roughly the same as that for the clean samples, with no significant deviation.

We further employ BIRD~\cite{DBLP:conf/nips/Chen000S23} to evaluate VDN, QMIX, and MAPPO models injected with BLAST across distinct maps (``8m'', ``3m'', and ``2s3z'') in the StarCraft2Env. Three maps are configured with specific triggers to generate backdoor models: $\mathcal{T}_1$ is deployed on ``8m'' and ``2s3z'', while $\mathcal{T}_2$ is applied to ``3m''. For each backdoor model, we use BIRD with an initial perturbation magnitude of 0.1 to generate a ``pseudo-trigger''. Then, these backdoor models are run to interact with the environment with and without pseudo-triggers, respectively. BIRD will use the maximum consecutive reward drop plus its standard deviation during those interactions without pseudo-triggers as the backdoor detection threshold to detect backdoor attacks.
The detection results show that in the ``8m'' map, BIRD fails to detect backdoor attacks under the VDN, QMIX, and MAPPO algorithms; in the ``3m'' map, BIRD only succeeds under VDN, but still fails under QMIX and MAPPO; and in the ``2s3z'' map, BIRD only succeeds under MAPPO, but fails under VDN and QMIX. The core method of BIRD's backdoor detection is to compare the abnormally large reward drop during the pseudo-trigger occurs with the reward changes when there is no pseudo-trigger. Since our BLAST attack delays the malicious action generation until after the backdoor is triggered, rather than being generated during the trigger occurs like existing attacks, the reward does not exhibit anomalous changes in the presence of pseudo triggers.

Based on the above results, we can find that our BLAST attack is sustainable against existing advanced defenses. The key reason for the detection failure is that our attack has a lag relative to the trigger, that is, the attack occurs after the trigger appears rather than immediately. This is different from the backdoor attacks in existing supervised learning and the instant backdoor attacks in DRL, whose triggers and attacks appear in pairs. One possible mechanism for detecting our kind of backdoor attacks is to identify anomalies in the temporal and spatial features of sequence data. This is still an open challenge.

\subsection{Ablation Studies}
\label{subsec:ablation}
Considering the impact of different parameters on the BLAST attack effect, we conduct ablation evaluations on the selection of different $\lambda$ in hacked reward and poisoning rate.

\subsubsection{$\lambda$ in Hacked Reward} 
\begin{figure}[t]
        \captionsetup{font=small}
	\centering
	\begin{subfigure}{0.49\linewidth}
		\centering
            \includegraphics[width=\linewidth]{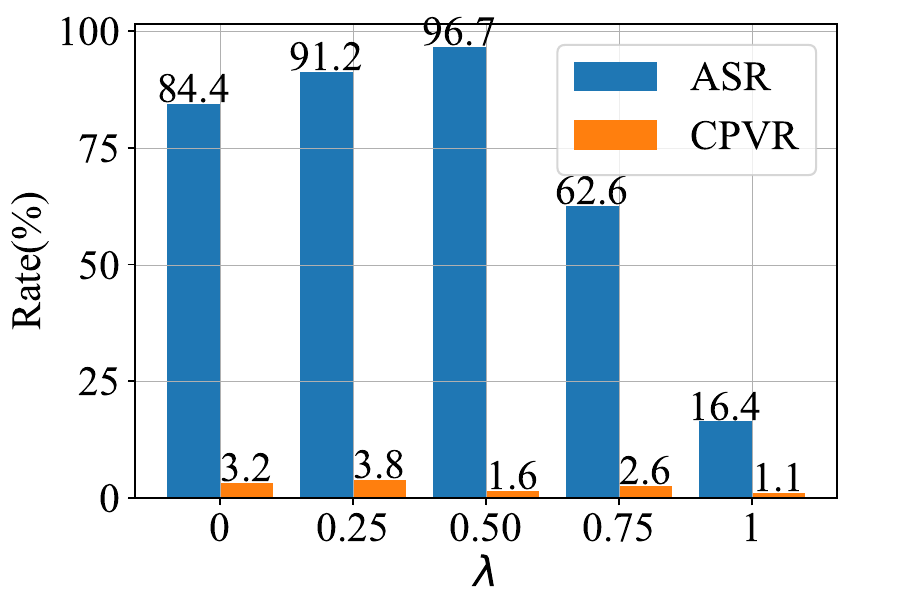}
		\caption{$\lambda$}
	\end{subfigure}
	\centering
	\begin{subfigure}{0.49\linewidth}
		\centering
	\includegraphics[width=\linewidth]{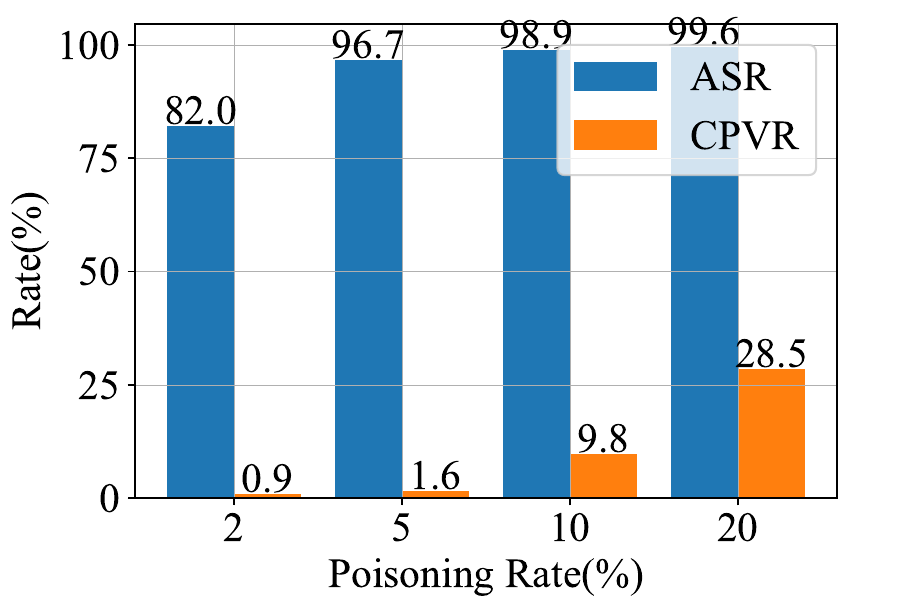}
		\caption{poisoning rates}
	\end{subfigure}
	\caption{ASR and CPVR achieved under different parameters.}
	\label{fig:asrcpvr}
\end{figure}

We further evaluate the performance of our BLAST backdoor attacks against VDN with different values of $\lambda \in \{0, 0.25, 0.50, 0.75, 1\}$ in hacked reward function, as shown in Fig \ref{fig:asrcpvr} (a).
We can find that when $\lambda = 0$, the $ASR$ can reach up to 84.4\%. This indicates that including only the first term in the hacked reward, \textit{i.e.}, target failure state guiding, is useful to some extent against c-MADRL. However, when $\lambda = 1$, the $ASR$ is only 16.4\%, although it can maintain a very low $CPVR$, which is 1.1\%. This suggests that the attack is not very effective when considering only the second term in the hacked reward, \textit{i.e.}, the effect of the BLAST agent on the next time-step actions of its teammates. Attacks perform best when $\lambda = 0.5$, being able to maintain a 1.6\% $CPVR$ and achieve a 96.7\% $ASR$. This further demonstrates the effectiveness of our hacked reward for backdoor attacks against c-MADRL.

\subsubsection{Poisoning Rate $pr$} We further evaluate the performance of our BLAST backdoor attacks with different values of poisoning rate $pr \in \{2\%, 5\%, 10\%, 20\% \}$, as shown in Fig. \ref{fig:asrcpvr} (b).
We can observe that as the poisoning rate increases, $ASR$ continues to rise, but $CPVR$ also continues to increase.
When the poisoning rate is 5\%, $ASR$ and $CPVR$ are a good balance point, which can make the attack effective while ensuring good concealment.

\section{Related Work}
\label{sec:related}
\begin{table*}[h!]
    \captionsetup{font=small}
    \centering
    \caption{Comparison of existing backdoor attacks against DRL. *-E denotes the type of environment targeted by the backdoor attack. *-T denotes the pattern of the backdoor trigger. Specifically, SA-E: Single-agent environment. TAC-E: Two-agent competitive environment. MAC-E: Multi-agent cooperative environment. I-T: Instant trigger. S-T: Spatial trigger. T-T: Temporal trigger. ST-T: Spatiotemporal trigger. OOD-T: Out-of-distribution trigger. ID-T: In-distribution trigger.}
    \label{tab:related work}
    \begin{tabular}{c|ccc|cccc|cc|c|c}
        \hline
        \multicolumn{1}{c|}{\multirow{2}{*}{Related work}} & \multicolumn{3}{c|}{Environment} & \multicolumn{4}{c|}{Trigger pattern} & \multicolumn{2}{l|}{Trigger source} & \multirow{2}{*}{Attack duration} & \multirow{2}{*}{Hacking method} \\\cline{2-10}
\multicolumn{1}{c|}{}                           & SA-E     & TAC-E     & MAC-E    & I-T     & T-T    & S-T    & ST-T    & OOD-T             & ID-T            &                                  &                                 \\\hline
        Kiourti \textit{et al.} \cite{9218663} & \ding{51} & \ding{55} & \ding{55} & \ding{51} & \ding{55} & \ding{55} & \ding{55} & \ding{51} & \ding{55} & Instant & Maximise target action reward\\
        Yang \textit{et al.} \cite{yang2019design} & \ding{51} & \ding{55} & \ding{55} & \ding{51} & \ding{55} & \ding{55} & \ding{55} & \ding{51} & \ding{55} & Persistent & Reverse reward\\
        Ashcraft \textit{et al.} \cite{ashcraft2021poisoning} & \ding{51} & \ding{55} & \ding{55} & \ding{51} & \ding{55} & \ding{55} & \ding{55} & \ding{55} & \ding{51} & Instant & Reverse reward\\\hline
        Wang \textit{et al.} \cite{STOPANDGO} & \ding{51} & \ding{55} & \ding{55} & \ding{51} & \ding{51} & \ding{55} & \ding{55} & \ding{55} & \ding{51} & Persistent & Specific reward\\
        Cui \textit{et al.} \cite{CuiHMJZ24} & \ding{51} & \ding{55} & \ding{55} & \ding{51} & \ding{55} & \ding{55} & \ding{55} & \ding{51} & \ding{55} & Instant & Maximise target action reward\\
        Chen \textit{et al.} \cite{BAFFLE} & \ding{51} & \ding{55} & \ding{55} & \ding{51} & \ding{55} & \ding{55} & \ding{55} & \ding{51} & \ding{55} & Instant & Reverse reward\\
        Yu \textit{et al.} \cite{yu2022temporal} & \ding{51} & \ding{55} & \ding{55} & \ding{55} & \ding{55} & \ding{51} & \ding{55} & \ding{55} & \ding{51} & Controllable & Reverse reward\\
        Yu \textit{et al.} \cite{yu2023spatiotemporal} & \ding{51} & \ding{55} & \ding{55} & \ding{55} & \ding{55} & \ding{55} & \ding{51} & \ding{55} & \ding{51} & Controllable & Specific reward\\\hline
        Wang \textit{et al.} \cite{ijcai2021p509} & \ding{55} & \ding{51} & \ding{55} & \ding{55} & \ding{55} & \ding{51} & \ding{55} & \ding{55} & \ding{51} & Persistent & Reverse reward\\\hline
        Chen \textit{et al.} \cite{MARNet} & \ding{55} & \ding{55} & \ding{51} & \ding{51} & \ding{55} & \ding{55} & \ding{55} & \ding{51} & \ding{51} & Instant & Specific reward\\
        Chen \textit{et al.} \cite{chen2022backdoor} & \ding{55} & \ding{55} & \ding{51} & \ding{51} & \ding{55} & \ding{55} & \ding{55} & \ding{51} & \ding{55} & Persistent & Reverse reward\\
        Zheng \textit{et al.} \cite{zheng2023one4all} & \ding{55} & \ding{55} & \ding{51} & \ding{51} & \ding{51} & \ding{55} & \ding{55} & \ding{55} & \ding{51} & Instant & Maximise target action reward\\
        Our \textbf{BLAST} & \ding{55} & \ding{55} & \ding{51} & \ding{55} & \ding{55} & \ding{55} & \ding{51} & \ding{55} & \ding{51} & Controllable & Specific reward\\
        \hline 
    \end{tabular}
    \vspace*{-3mm} 
\end{table*}
\subsection{Backdoor Attack against DRL}
Most existing studies on backdoor attacks focus on DNN \cite{DBLP:journals/tnn/LiJLX24}.
Due to the representation ability and feature extraction ability inherited from DNN, DRL also inevitably faces this threat.
Kiourti \textit{et al.} \cite{9218663} propose a backdoor attack method against DRL with image patch triggers. During training, when the model's input contains a trigger and the output is the target action or a random one, they maximize the corresponding reward value to achieve a targeted or untargeted attack.
Yang \textit{et al.} \cite{yang2019design} investigate the persistent impact of backdoor attacks against the LSTM-based PPO algorithm \cite{schulman2017proximal}. When an image patch trigger appears at a certain time step, the backdoor agent persistently goes for the attacker-specified goal instead of the original goal.
Ashcraft \textit{et al.} \cite{ashcraft2021poisoning} use in-distribution triggers and multitask learning to train a backdoor agent. An in-distribution trigger is an observation that is not anomalous to the environment and is therefore more difficult to detect.
Wang \textit{et al.} \cite{ijcai2021p509} study backdoor attacks against DRL in two-player competitive games. The opponent's actions are used as a trigger to switch the backdoor agent to fail fast.
Wang \textit{et al.} \cite{STOPANDGO} study backdoor attacks on DRL-based traffic congestion control systems and used a specific set of combinations of positions and speeds of vehicles in the observation as the trigger. When the trigger is present, the backdoor DRL model could generate malicious decelerations to cause a physical crash or traffic congestion.
Cui \textit{et al.} \cite{CuiHMJZ24} propose a sparse backdoor attack that selects observations with high attack values to inject triggers and generates trigger patterns using mutual information-based tuning.
Chen \textit{et al.} \cite{BAFFLE} study backdoor attacks against offline reinforcement learning, which poison open-source reinforcement learning datasets to implant backdoors to DRL agents. They use instant triggers and explore the effectiveness of attacks with a distributed strategy and a one-time strategy.
Our previous works \cite{yu2022temporal, yu2023spatiotemporal} studied a new temporal and spatial-temporal backdoor trigger to DRL, which has shown high concealment.

\subsection{Backdoor Attack against c-MADRL}
So far, only a few studies have been done on backdoors in c-MADRL. Chen \textit{et al.} \cite{MARNet} study backdoor attacks against c-MADRL. They consider both out-of-distribution and in-distribution triggers and use a pre-trained expert model to guide the selection of actions of poisoned agents. Besides, they modify the team reward to encourage poisoned agents to exhibit their worst actions. This method implants backdoors in all agents, and all agents will perform an instant malicious action when observing the trigger.
Chen \textit{et al.} \cite{chen2022backdoor} propose a backdoor attack that affects the entire multi-agent team by targeting just one backdoor agent. This attack requires a random network distillation module and a trigger policy network to guide the backdoor agent on what actions to select and when to trigger the backdoor. During execution, the backdoor agent needs to rely on the trigger policy network to determine whether to trigger the backdoor.
Zheng \textit{et al.} \cite{zheng2023one4all} also implant a backdoor into one of the agents in a multi-agent team, and use distance as a condition to trigger the backdoor. When the trigger condition is met, the backdoor agent performs a specified action which leads to the failure of the team task.
These existing backdoor attacks are effective against c-MADRL, but they either poison all agents, require additional networks to activate backdoors, or do not take into account the influence between agents. We compare these existing backdoor attacks against DRL in Table \ref{tab:related work}.

\subsection{Backdoor Defense Mechanism}
Existing backdoor defenses broadly include backdoor detection, data filtering, and backdoor mitigation.
\cite{SpectralSignatures,chen2019ac} are two poisoned input detection methods that depend on the difference of covariance spectrum of the feature representation and activation value distributions in the last hidden neural layer between poisonous and clean samples, respectively.
Wang \textit{et al.} \cite{DBLP:conf/sp/WangYSLVZZ19} use reverse engineering to infer and detect the shapes and locations of backdoor triggers and mitigate backdoors through input filters, neuron pruning, and unlearning techniques.
Neuralsanitizer \cite{neuralsanitizer2024zhu} is a backdoor detection and removal tool that identifies real triggers and detects backdoors by reconstructing a set of potential triggers and removing redundant features.
For backdoor attacks against DRL, PolicyCleanse \cite{DBLP:conf/iccv/GuoLW023} is a backdoor detection and mitigation approach that uses policy optimization with a reversed cumulative reward of the backdoor agent in competitive DRL policies. Sanyam \textit{et al.} \cite{DBLP:conf/sp/VyasHM24} develop a classifier that utilizes the neuron activation patterns in DRL to detect the presence of in-distribution backdoor triggers. Both BIRD \cite{DBLP:conf/nips/Chen000S23} and SHINE\cite{yuan2024shine} claim to detect backdoor policies both in single- and multi-agent DRL scenarios. BIRD requires accessing the clean and poisoned environments to compare reward change behaviors and restore the potential trigger. SHINE collects a set of poisoned trajectories to identify features in the trigger-presenting states leading to backdoor actions and deems them as the backdoor trigger. All existing defense methods assume that backdoor actions are immediately generated when the trigger occurs. Since BLAST is hidden in only one of multiple DRL agents, whose triggers and actions are decoupled and distributed in different time series, rather than in individual input, most of the existing methods will fail to detect this attack-trigger-decoupled attack.

\section{Conclusions and Future}
\label{sec:con}
In this paper, we study backdoors against c-MADRL. To enhance the stealthiness, effectiveness, and practicality of backdoor attacks, we propose a novel backdoor attack against c-MADRL, BLAST, which can disrupt the entire multi-agent team by implanting a backdoor in only one agent. We use spatiotemporal features rather than an instant state as the backdoor trigger and design the malicious reward function based on unilateral influence. We evaluate our BLAST against VDN, QMIX, and MAPPO algorithms in two c-MADRL environments and three backdoor defense methods. The experimental results demonstrate that BLAST can achieve a high attack success rate and a low clean performance variance rate. In the future, we will explore backdoor attacks in black-box scenarios, as well as study effective defense methods for c-MADRL backdoors.

\bibliographystyle{IEEEtran}
\bibliography{IEEEabrv, ref}
\end{document}